%% file: _main.tex
\documentclass[letterpaper]{article} 
\usepackage{aaai2026}  
\usepackage{times}  
\usepackage{helvet}  
\usepackage{courier}  
\usepackage[hyphens]{url}  
\usepackage{graphicx} 
\usepackage{natbib}  
\usepackage{caption} 
\usepackage{algorithm}
\usepackage{algorithmic}

\usepackage{newfloat}
\usepackage{listings}
\DeclareCaptionStyle{ruled}{labelfont=normalfont,labelsep=colon,strut=off} 
\floatstyle{ruled}
\newfloat{listing}{tb}{lst}{}
\floatname{listing}{Listing}
\nocopyright

\usepackage{booktabs}
\usepackage{multirow}
\usepackage{xcolor}
\usepackage{arydshln}
\usepackage{makecell}
\usepackage{amsmath}
\usepackage{amssymb}
\usepackage{mathtools}
\usepackage{amsthm}
\usepackage{subcaption}

\definecolor{darkergreen}{RGB}{21, 152, 56}
\definecolor{red2}{RGB}{252, 54, 65}
\definecolor{grey}{RGB}{165, 165, 165}
\usepackage{pifont}
\newcommand{\cmark}{\ding{51}}
\newcommand{\xmark}{\textcolor{grey}{\ding{55}}}

\usepackage{xspace}
\newcommand{\scii}{StarCraft II\xspace}
\newcommand{\benchmark}{SC2Arena\xspace}
\newcommand{\method}{StarEvolve\xspace}

\usepackage{mdframed}
\usepackage[many]{tcolorbox}  
 
\definecolor{swecream}{RGB}{255,247,236}
\newtcolorbox{promptbox}[1]{
    enhanced,
    breakable,
    boxrule=1pt,  
    fontupper=\small,
    fonttitle=\bfseries\color{black},
    arc=3pt,  
    rounded corners,
    colframe=black,
    colbacktitle=swecream,
    colback=swecream,
    title=#1,
    left=2mm,  
    right=2mm,  
    top=1mm,  
    bottom=1mm  
}

\title{SC2Arena and StarEvolve: Benchmark and Self-Improvement \\ Framework for LLMs in Complex Decision-Making Tasks}
\author{
    Pengbo Shen\textsuperscript{\rm 1}\equalcontrib, 
    Yaqing Wang\textsuperscript{\rm 1}\equalcontrib, 
    Ni Mu\textsuperscript{\rm 2}\equalcontrib, 
    Yao Luan\textsuperscript{\rm 2}, 
    Runpeng Xie\textsuperscript{\rm 1}, 
    Senhao Yang\textsuperscript{\rm 1}, 
    Lexiang Wang\textsuperscript{\rm 1}, 
    Hao Hu\textsuperscript{\rm 3}, 
    Shuang Xu\textsuperscript{\rm 1}\thanks{Corresponding authors.}, 
    Yiqin Yang\textsuperscript{\rm 1}\footnotemark[2], 
    Bo Xu\textsuperscript{\rm 1}\footnotemark[2]
}
\affiliations{
    $^1$
    The Key Laboratory of Cognition and Decision Intelligence for Complex Systems,\\Institute of Automation, Chinese Academy of Sciences\\
    $^2$Beijing Key Laboratory of Embodied Intelligence Systems, \\ Department of Automation, Tsinghua University\\ 
    $^3$Moonshot AI, Beijing, China \\
    pengbo.shen@outlook.com, wangyaqing2024@ia.ac.cn, mn23@mails.tsinghua.edu.cn
}

\isChecklistMainFile

\begin{document}

\maketitle

\input{text/0_abstract}

\input{text/1_intro}

\input{text/2_related_work}

\input{text/3_benchmark}

\input{text/4_method}

\input{text/5_experiment}

\input{text/6_conclusion}


\bibliography{_ref}

\onecolumn

\appendix

\input{appendix/benchmark_detail}

\input{appendix/method_detail}

\input{appendix/experiment_detail}

\input{appendix/visualization_detail}

\input{appendix/additional_experiments}

\end{document}

%% file: text/0_abstract.tex
\begin{abstract}

Evaluating large language models (LLMs) in complex decision-making is essential for advancing AI's ability for strategic planning and real-time adaptation. 
However, existing benchmarks for tasks like StarCraft II fail to capture the game's full complexity, such as its complete game context, diverse action spaces, and all playable races. 
To address this gap, we present SC2Arena, a benchmark that fully supports all playable races, low-level action spaces, and optimizes text-based observations to tackle spatial reasoning challenges.
Complementing this, we introduce StarEvolve, a hierarchical framework that integrates strategic planning with tactical execution, featuring iterative self-correction and continuous improvement via fine-tuning on high-quality gameplay data.
Its key components include a Planner-Executor-Verifier structure to break down gameplay, and a scoring system for selecting high-quality training samples.
Comprehensive analysis using SC2Arena provides valuable insights into developing generalist agents that were not possible with previous benchmarks. 
Experimental results also demonstrate that our proposed StarEvolve achieves superior performance in strategic planning.
Our code, environment, and algorithms are publicly available.

\end{abstract}


%% file: text/1_intro.tex
\section{Introduction}
\label{sec:intro}

\begin{figure*}[t]
    \centering
    \begin{minipage}{0.98\textwidth}
        \centering
        \includegraphics[width=\textwidth]{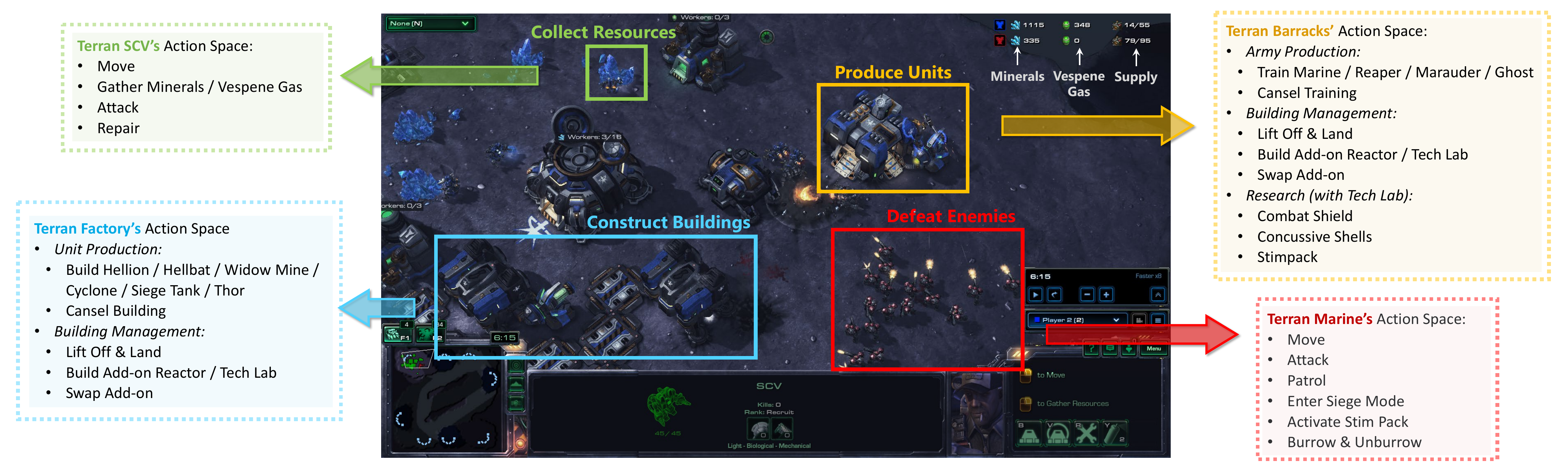}
    \end{minipage} 
    \\
    \vspace{4pt}
    \begin{minipage}{0.98\textwidth}
        \centering
        \includegraphics[width=\textwidth]{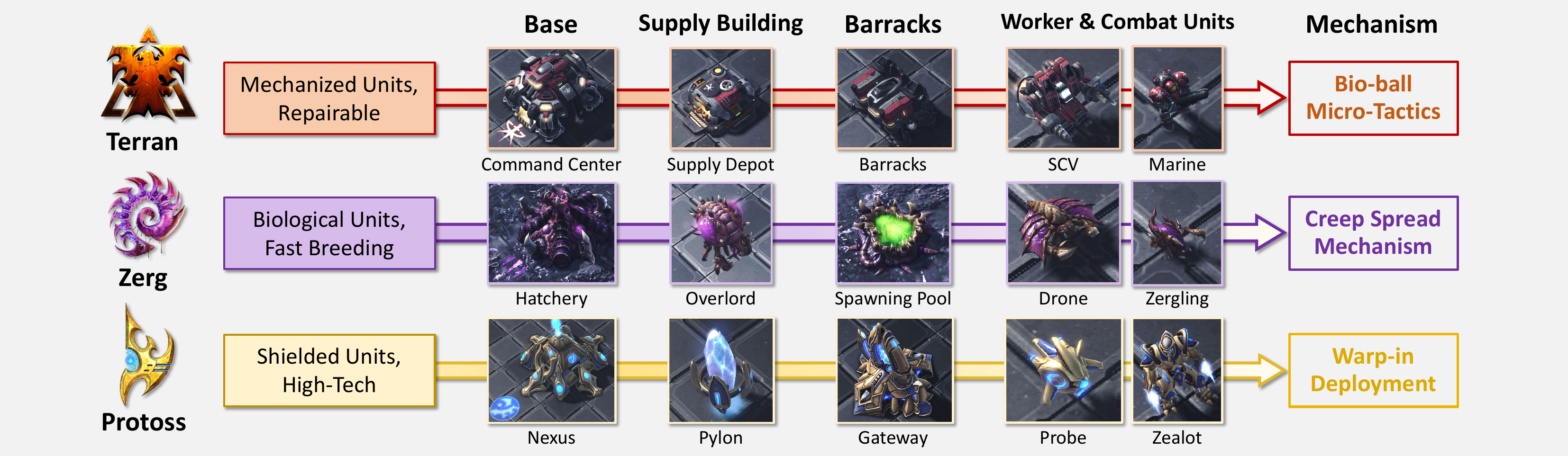}
    \end{minipage}
    \caption{An illustration of \scii. 
\scii is a highly complex and competitive real-time strategy game, where players choose one of three distinct races (Terran, Zerg, Protoss) to gather resources, construct buildings, produce units, and defeat enemies. 
Each race has unique traits and can produce race-specific buildings and military units.}
    \label{fig:sc2_illustration}
\end{figure*}

The rapid advancement of large language models (LLMs) \cite{achiam2023gpt, guo2025deepseek} has accelerated research towards artificial general intelligence \cite{mon2025embodied, zheng2025large}. A critical milestone in this pursuit is the development of agents capable of strategic planning and real-time adaptation in complex environments.
\scii, a renowned real-time strategy game, provides an ideal testbed with its profound strategic depth, diverse unit types and interactions, vast decision space, and inherent demands for world modeling, as illustrated in Fig. \ref{fig:sc2_illustration}. 
These characteristics align closely with the LLMs' core strengths \cite{wei2022chain, yao2023react, wang2023plan}, making it particularly suitable for LLM evaluation.
Therefore, establishing benchmarks in \scii to assess LLM performance is critical.
However, existing LLM benchmarks in \scii \cite{ma2024large, li2024pysc2} often fail to capture the original game's complexity, lacking support for full-length game context, complete low-level action spaces, comprehensive race configuration, and agent-vs-agent mechanisms. 
This limits their ability to evaluate the strategic reasoning and adaptability of LLM agents. 



To bridge this gap, we propose \textbf{\benchmark}, a comprehensive \scii benchmark for evaluating LLMs' decision-making ability. 
With a standardized JSON interface, it supports \ding{172} full-length game context, \ding{173} all playable races (Terran, Protoss, Zerg), and \ding{174} complete low-level actions, preserving the game's strategic depth. 
Crucially, \benchmark addresses inherent challenges of text-based agents by optimizing observations: proximity-based unit ordering captures units' spatial relationships, and worker aggregation reduces information overload. 
Additionally, it enables agent-vs-agent gameplay, facilitating direct competition and an agent ranking mechanism to better evaluate performance and strategic adaptability.  

Furthermore, we introduce \textbf{\method}, a closed-loop agent framework that enables self-correction and self-improvement for strategic decision-making. 
It employs a hierarchical architecture: a Planner generates high-level strategic commands, while an Executor converts them into precise low-level actions. 
Both components integrate Verifier modules that perform iterative self-correction to ensure syntactic correction and feasibility. 
Crucially, \method achieves continuous self-improvement by collecting gameplay data and performing supervised fine-tuning (SFT). 
We design an RL-inspired scoring function to prioritize high-quality training samples, which quantifies action impact through discounted value increments across key game-state metrics. 
Utilizing these data, we fine-tune the Planner, Executor, and Verifiers to continuously improve their performance.






In summary, our contributions are twofold: 
(1) We propose \benchmark, a comprehensive benchmark for evaluating LLMs in \scii, supporting full-length game context, all races, full action spaces, and agent-vs-agent gameplay.
(2) We propose \method, a closed-loop agent framework, which integrates strategic planning with tactical execution. It achieves self-correction with iterative decision refinement, and also continuous self-improvement via SFT, enabling LLM agents to defeat challenging opponents.







%% file: text/2_related_work.tex
\section{Related Work}

\noindent \textbf{LLM as autonomous agents. }
Recent advances in LLMs have empowered autonomous agents to succeed in diverse domains, including conversational games \cite{xu2023exploring, wang2023avalon}, webpage navigation \cite{lai2024autowebglm}, and open-world environments \cite{wang2023voyager}. 
Concurrently, benchmarks \cite{liu2023agentbench, zhou2023webarena, light2023avalonbench, rismanchian2024turtlebench} have emerged to evaluate these agents' planning and reasoning capabilities. 
Despite these successes, LLM agents remain relatively underexplored in real-time adversarial games, which require integration of long-term strategic planning and precise tactical execution, under partial observability and time constraints. 

\vspace{3pt}
\noindent \textbf{AI in \scii. }
\scii remains a classical AI research environment \cite{ontanon2013survey} due to its expansive action space and real-time decision complexity. 
Prior work primarily leveraged reinforcement learning (RL), achieving strong micro-management performance through multi-agent RL \cite{samvelyan19smac, rashid2020monotonic, yu2022surprising}. 
Subsequent efforts like \citep{vinyals2019grandmaster} scaled RL to full-length game contexts, although this approach required massive training resources and carefully engineered rewards. 
In contrast, LLMs offer complementary strengths to strategic planning by leveraging inherent world knowledge and reasoning capabilities \cite{yildirim2024task, yuan2024advancing}. 
Yet, current benchmarks for evaluating LLM decision-making in \scii \cite{ma2024large, li2024pysc2} have limitations, including the abstraction of low-level tactical execution, a lack of support for full-length game contexts and full race configurations, and insufficient agent-vs-agent mechanisms. 
To bridge these gaps, we propose \benchmark, a comprehensive \scii benchmark to evaluate agents across diverse dimensions. 
We further propose \method, a self-improving agent framework for this benchmark, advancing LLM-driven decision-making in complex environments.

%% file: text/3_benchmark.tex
\begin{figure*} [t]
    \centering
    \includegraphics[width=0.8\linewidth]{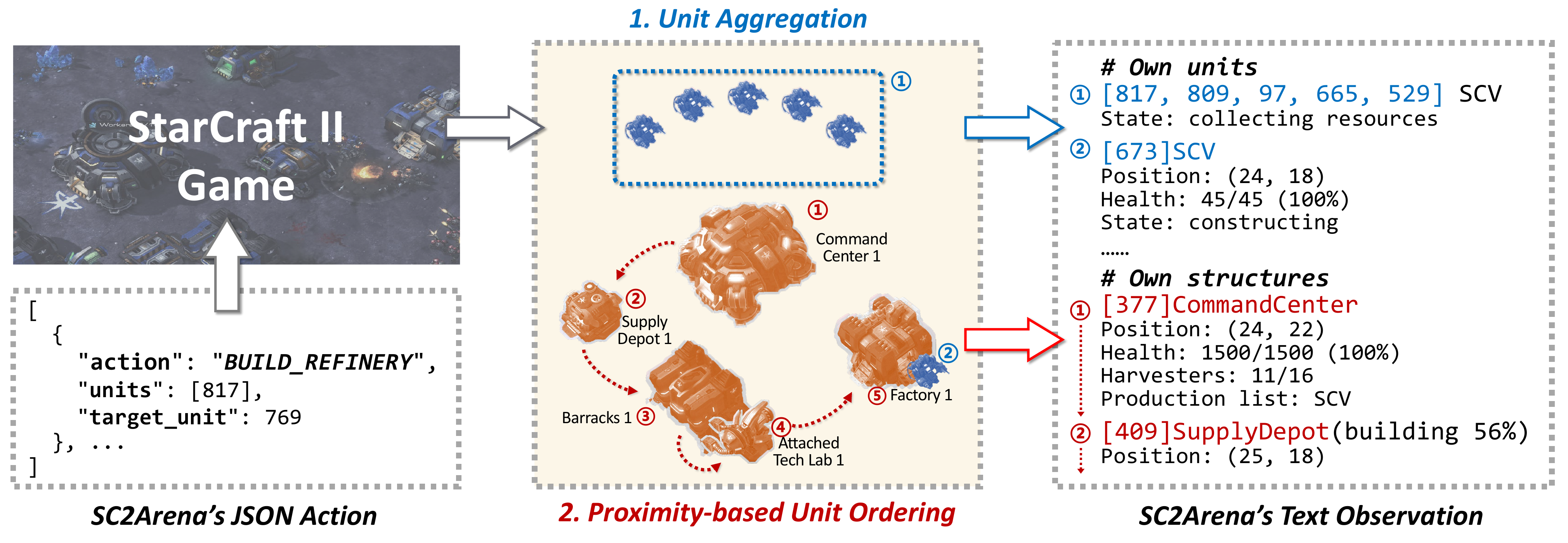}
    \caption{Illustration of \benchmark design. 
    (a) SC2Arena provides a standardized and extensible text-based interface where actions are defined in a unified JSON format. 
    (b) To convert game states into textual observations suitable for LLMs, SC2Arena introduces two key techniques: Unit aggregation and Proximity-based unit ordering.}
    \label{fig:benchmark_design}
\end{figure*}

\input{table/benchmark_comparison_flatten}




\section{Benchmark: \benchmark}
\label{sec:benchmark}

In this section, we propose \benchmark, a structured text-based benchmark designed to evaluate LLM agents in \scii, with standardized observation and action interfaces. 
\benchmark supports full-length game contexts, complete low-level action spaces, all three races and agent-vs-agent gameplay, thus preserving the original game's strategic depth and complexity.

\subsection{Introduction to \benchmark}
\label{subsec:benchmark_intro}

\noindent \textbf{Overview of \scii. }
\scii is a highly complex and competitive real-time strategy game, where players control one of three distinct races (Terran, Zerg, Protoss) to gather resources, construct buildings, produce units, and command armies to defeat opponents. 
As shown in Fig. \ref{fig:sc2_illustration}, the game demands proficiency in macro-management, like resource management and base-building, micro-level unit control, and strategic decision-making, all under real-time constraints and partial information. 
Its observation space can reach up to 1,200 dimensions, while the action space can extend to 300 dimensions. 
Professional matches typically last 20--30 minutes, with players executing hundreds of actions per minute. 
This complexity makes it both a popular AI benchmark and a globally renowned e-sport.

Each race in \scii offers a distinct playstyle and unique mechanics, as shown in Fig. \ref{fig:sc2_illustration}:
\textbf{Terrans} are a human race with mechanized, repairable units. 
Their unique Bio-ball Micro-Tactic involves tightly grouped armies supported by Medivacs (flying medical transport units). 
By optimizing unit positioning and treating injured units, this strategy maximizes firepower and survivability.
\textbf{Zerg} are a swarm-based alien race with rapidly breeding biological units to overwhelm opponents. 
Their unique Creep Spread Mechanism enhances unit speed and vision, enabling aggressive expansion and superior map control. 
\textbf{Protoss} are an advanced alien race with regenerating shielded units. Their unique Warp-in Deployment mechanic allows armies to instantly arrive at strategic locations within their Pylon network, enabling precise timing attacks. 
Appendix \ref{app:scii_intro} offers further details.

\vspace{3pt}
\noindent \textbf{Improvements over prior benchmarks. }
As shown in Table \ref{tab:benchmark_comparison}, while prior benchmarks mainly focus on micro-management scenarios \cite{samvelyan19smac, li2024pysc2} or high-level strategic decision-making \cite{ma2024large}, our benchmark, \benchmark, offers a more comprehensive evaluation platform covering full-length game contexts that combine strategic planning and tactical execution. 
Unlike existing LLM-agent benchmarks \cite{ma2024large}, \benchmark supports agent-vs-agent gameplay, enabling mechanisms such as self-play and agent ranking to advance competitive AI development. 
Moreover, compared to \citet{ma2024large}, \benchmark supports all three races, enables gameplay on any map, and offers precise control over the complete low-level action space, preserving the game's original complexity and richness.

\begin{figure*}[t]
    \centering
    \includegraphics[width=0.98\linewidth]{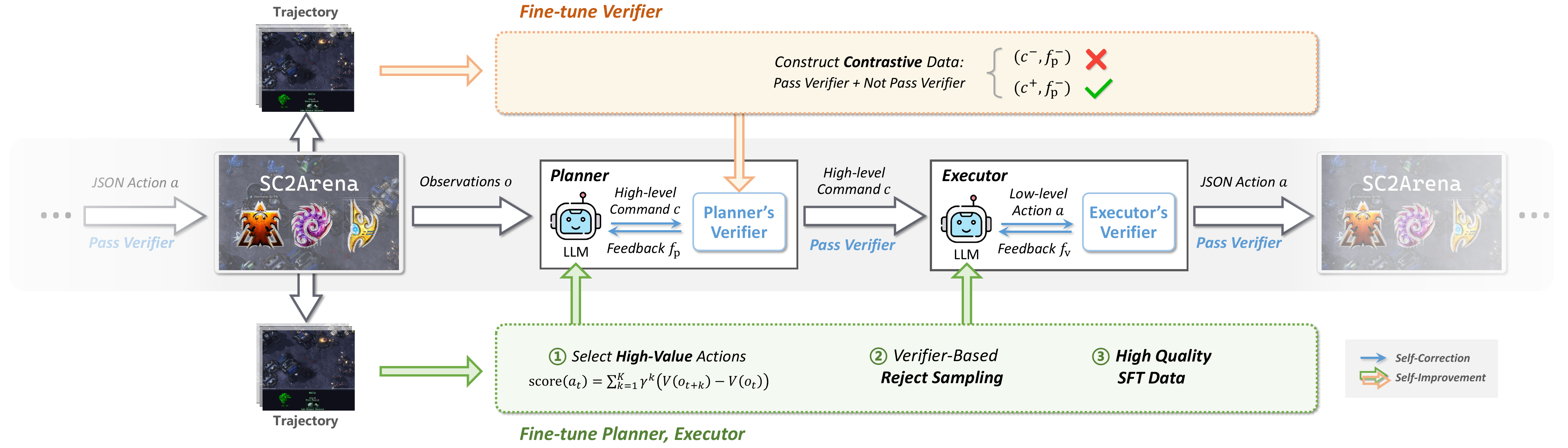}
    \caption{Illustration of \method.
    The Planner produces strategic commands, while the Executor generates executable actions, both incorporating Verifiers to enable iterative self-correction. 
    Furthermore, \method achieves self-improvement by collecting gameplay trajectories, selecting high-quality or contrastive samples, and employing supervised fine-tuning. 
    }
    \label{fig:agent_illustration}
\end{figure*}

\subsection{Benchmark Design}
\label{subsec:benchmark_trick}

To enable LLM agents to play \scii effectively, game states must be converted into textual observations, which presents two key challenges:
(1) \textbf{\textit{Spatial reasoning difficulty}}: Agents struggle to infer positional relationships between units based solely on text \cite{sharma2023exploring}. 
This hinders agents' ability to identify attached and interactable unit structures, such as adjacent Barracks and Tech Labs, thereby reducing their effectiveness. 
(2) \textbf{\textit{Information overload}}: Raw observations often include excessive low-value data, such as repetitive worker unit statuses, which overwhelms agents and distracts them from strategic decision-making  \cite{adejumo2024towards}.

To address these issues, \benchmark introduces two key techniques, as illustrated in Fig. \ref{fig:benchmark_design}:
(1) \textbf{\textit{Proximity-based unit ordering}}: 
We enhance spatial reasoning via a greedy nearest-neighbor unit ordering, which explicitly encodes spatial relationships into the observation text. 
Specifically, starting from the Command Center, each unit is listed after its nearest neighbor unit. 
This heuristic helps agents identify interactable structures effectively. 
(2) \textbf{\textit{Unit aggregation}}: 
We reduce information overload by aggregating routine worker units into a single representation, preserving core workers' information while removing redundant details. 
As validated in Section \ref{subsec:ablation_study}, these techniques improve agent performance.

\vspace{3pt}
\noindent \textbf{Observation and Action Interface. }
As shown in Fig. \ref{fig:benchmark_design}, \benchmark provides a standardized and extensible text-based interface for \scii. 
Observations are formatted as a hierarchical text summary, categorized into six key types: resource levels, unit states, building statuses, ongoing activities, visible enemy units/structures, and technological progress. 
Actions are defined in a unified JSON format, enabling agents to specify tasks such as unit production, structure construction, and tactical maneuvers. 
Each action includes parameters like unit IDs, target locations, or specific abilities, allowing precise in-game control. 
Appendix \ref{app:obs_example} provides detailed examples of observations and actions. 


\vspace{3pt}
\noindent \textbf{Implementation details. }
At each iteration of the gameplay loop, the agent processes the textual game state, constructs a prompt, and queries the LLM for a decision. The environment extracts the valid JSON from the LLM's output, parses it into an executable action, and executes it. This design maintains a clear separation between the agent and the environment. Appendix \ref{app:benchmark_implementation} provides more details. 




\subsection{Evaluation Metrics}
\label{subsec:benchmark_metrics}

\benchmark supports competitive interactions between agents and various opponents, including built-in AI, rule-based adversaries, or other agents. We provide a comprehensive set of metrics to assess both performance and efficiency across these competitive settings.

\vspace{3pt}
\noindent \textbf{\textit{Performance metrics:}} 
(1) Elo Rating (ELO): A widely adopted metric in competitive environments for quantifying agents' relative levels. It is derived from agent-vs-agent match outcomes and updates after each contest. 
(2) Win Rate (WR): Percentage of games won, typically against built-in AIs. 
(3) Time Cost of Winning (TCW): Average duration of winning games, reflecting strategic efficiency. 
(4) Supply Block Ratio (SBR): Percentage of game time during which a player cannot produce new units due to reaching the supply cap. This metric evaluates a player's ability to manage resources and sustain continuous military production.
(5) Resource Utilization Ratio (RUR): Average minerals and gas spent per game timestep, reflecting economic development. 

\vspace{3pt}
\noindent \textbf{\textit{Efficiency metrics:}}
(6) Tokens Per Decision (TPD): Average number of LLM tokens per decision.
(7) Valid Action Ratio (VAR): Percentage of valid and executable decisions made by the agent. 
Detailed formulations are provided in Appendix \ref{app:metrics}.

%% file: table/benchmark_comparison_flatten.tex
\begin{table*}[t]
\centering
\caption{Comparison of benchmarks in \scii from four dimensions.}
\label{tab:benchmark_comparison}
\begin{tabular}{lcccc}
\toprule
Benchmark & 
Agent-vs-agent & 
Low-level action space & 
All races & 
Full-length game context \\
\midrule
SMAC \cite{samvelyan19smac}
& \xmark & \cmark & \xmark & \xmark \\
LLM-PySC2 \cite{li2024pysc2}
& \xmark & \cmark & \xmark & \cmark \\
TextStarCraft II \cite{ma2024large}
& \xmark & \xmark & \xmark & \cmark \\
\midrule
\benchmark (ours)
& \cmark & \cmark & \cmark & \cmark \\
\bottomrule
\end{tabular}
\end{table*}

%% file: text/4_method.tex
\section{\method}
\label{sec:method}

In this section, we propose \method, a closed-loop, self-improving, and self-correcting LLM agent framework for \scii. 
\method comprises a high-level strategic Planner, a low-level Executor, and a Verifier mechanism for \textbf{self-correction}. 
Crucially, \method enhances its performance by generating high-quality gameplay data and fine-tuning itself, forming an effective \textbf{self-improvement} loop. 
This closed-loop design allows smaller LLMs to achieve competitive performance against challenging opponents.

\begin{figure*}[t]
    \centering
    \includegraphics[width=0.85\linewidth]{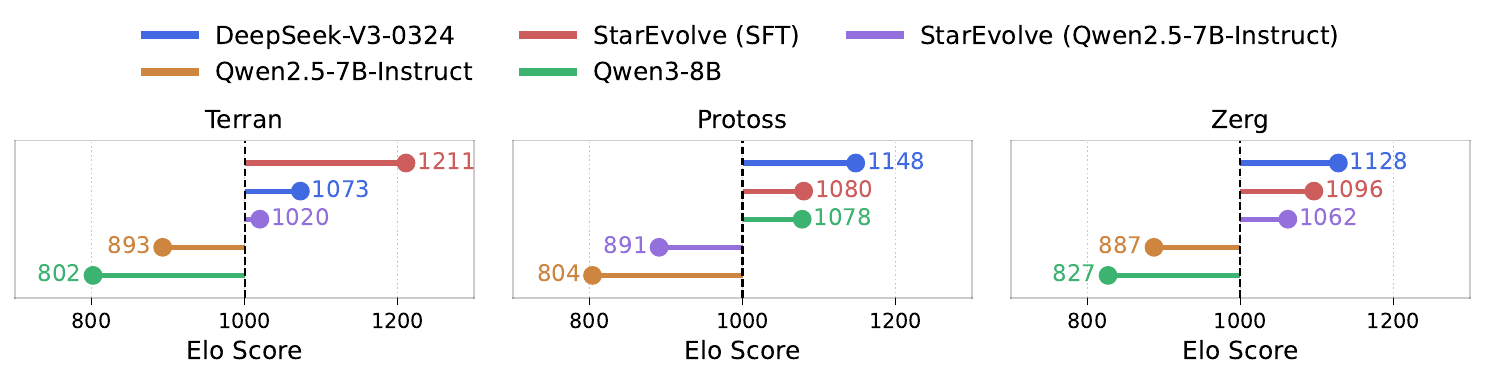}
    \caption{Elo rating results of \method and baseline agents. We report the average results of 20 Terran vs. Terran (TvT), 20 Protoss vs. Protoss (PvP), and 20 Zerg vs. Zerg games (ZvZ), respectively.}
    \label{fig:elo_results}
\end{figure*}

\input{table/main_results}

\subsection{Hierarchical Self-Correction Framework}
\label{subsec:self_correction}

Deploying LLM agents for \scii presents challenges, due to the game's vast low-level action space and the need for precise unit selection in JSON syntax, which often results in invalid actions, distracting agents from high-level strategy \cite{tam2024let}. 
To address this, \method employs a two-tier architecture: a high-level \textbf{\textit{Planner}} generates strategic commands, and a low-level \textbf{\textit{Executor}} translates them into JSON actions. 
However, the Planner may produce commands violating resource constraints or referencing invalid units, while the Executor can generate syntactically incorrect or infeasible actions. 
To mitigate these errors, \method integrates \textbf{\textit{Verifiers}} for both components, enabling iterative self-correction. 
The Planner's Verifier validates strategic commands against expert rules, while the Executor's Verifier ensures action syntax and feasibility. 
Fig. \ref{fig:agent_illustration} depicts this process.

\vspace{3pt}
\noindent \textbf{Planner. }
The Planner generates high-level strategic commands $c$ in natural language, such as ``Build a Barracks'', based on the game observations $o$ from \benchmark. 
To enhance command quality, the Planner's Verifier, also an LLM, evaluates each command. 
If any rule is violated, the Verifier offers structured feedback and suggestions $f_\text{p}$, allowing the Planner to iteratively refine its output. 
This design is highly extensible and supports seamless integration of additional expert knowledge. 
Appendix \ref{app:planner_prompt} provides detailed prompts and rule lists.

\vspace{3pt}
\noindent \textbf{Executor. }
The Executor translates the Planner's natural language commands $c$ into executable, low-level JSON actions $a$. 
To ensure the feasibility of actions, the Executor's Verifier validates each JSON output by checking (1) JSON syntax correction, (2) parameter and action validity, and (3) resource feasibility. 
Invalid actions also trigger structured feedback $f_\text{e}$, enabling iterative self-correction. 
Appendix \ref{app:executor_prompt} provides implementation details.

\begin{figure*}[t]
    \centering
    \includegraphics[width=\textwidth]{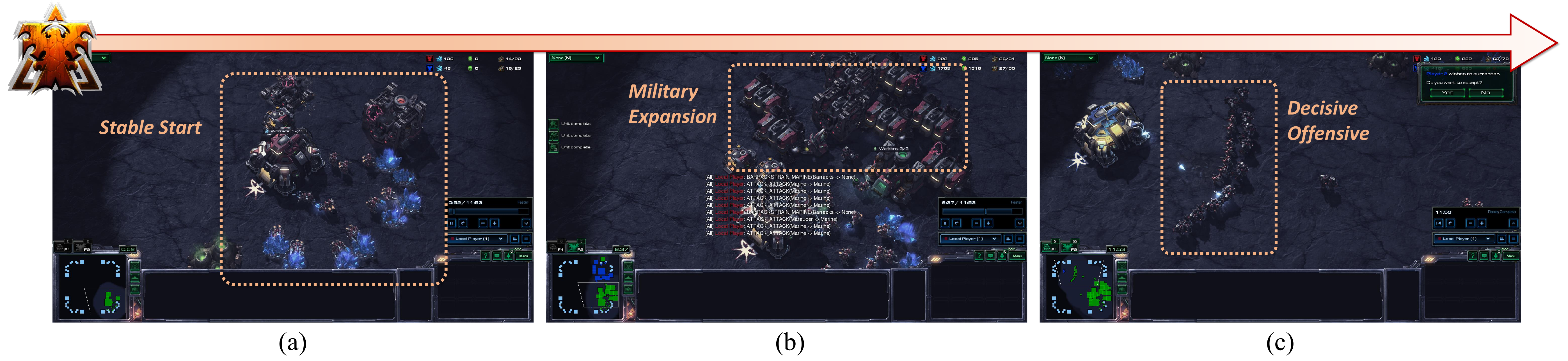}
    \caption{Visualizations of \method's gameplays. For the Terran vs. Terran game, \method transitions mid-game advantages into military strength and secures victory with a decisive offensive. More visualizations are presented in Appendix \ref{app:visualization}. }
    \label{fig:visualization_terran}
\end{figure*}

\subsection{Self-Improvement via Fine-tuning}
\label{subsec:self_improvement}

In this subsection, we design a self-improvement loop based on SFT as in Fig. \ref{fig:agent_illustration}, which enables the agent to extract strategic insights from its own high-performing behaviors and enhance its performance. 
We introduce the self-improvement loop from the following perspectives:

\vspace{3pt}
\noindent \textbf{SFT Data Construction.}
We construct the SFT dataset from winning game trajectories, where Chain-of-Thought (CoT) naturally emerges through in-game prompting. 
To ensure dataset quality, we design a scoring function inspired by reinforcement learning. 
Each action $a_t$ at timestep $t$ is evaluated based on its discounted future impact on core game metrics, including the converted mineral/vespene values of all units and structures, as well as the counts of army and mining units. 
Formally, the score aggregates temporally discounted gains over a 20-step horizon (30 seconds), where $M$ denotes the metric set and $V_m(o_t)$ represents the value of metric $m$ on the game observation $o_t$:
\begin{equation}
\text{score}(a_t) = \sum_{k=1}^{20} \gamma^k \sum_{m \in \mathcal{M}} \left( V_m(o_{t+k}) - V_m(o_t) \right), ~ \gamma = 0.95
\end{equation}  
To ensure comparability, raw metrics are normalized to zero mean and unit variance. 
Actions with scores above a predefined threshold are retained, prioritizing those that optimize both short-term tactical benefits and long-term strategic impact. 
Appendix \ref{app:metrics} provides the full evaluated metric list and more details. 

\vspace{3pt}
\noindent \textbf{Fine-tuning Planner and Executor. }
To enhance performance, we train the Planner and Executor with verifier-guided rejection sampling.
Specifically, the Planner is trained on $\langle$\textit{observation $o_t$, verified high-level command $c_t$}$\rangle$ pairs, while the Executor is trained on $\langle$(\textit{observation $o_t$, high-level command $c_t$}), \textit{verified JSON action $a_t$}$\rangle$ pairs. 
Crucially, the SFT loss is computed solely on the final output of each module's self-correction chain, masking all intermediate refinement steps. 
This approach mimics rejection sampling: the Verifier iteratively discards suboptimal outputs, accepting only the terminal responses as training targets, thereby learning the optimal decisions.

\vspace{3pt}
\noindent \textbf{Fine-tuning Planner's Verifier. }
To improve verification efficiency, we train the Planner's Verifier using self-correction trajectories. 
For each successfully verified high-level command $c^{+}$, we construct training data using the final accepted command and its immediate rejected predecessor $c^{-}$, in the form of $\langle$rejected command $c^{-}$ + feedback $f_\text{p}^{-}$ $\rightarrow$ accepted command $c^{+}$ + feedback $f_\text{p}^{+}$ $\rangle$. 
This method performs implicit contrastive learning, enabling the Verifier to better capture the boundary between valid and invalid commands compared to learning from isolated examples, thus enhancing learning effectiveness.

\vspace{3pt}
\noindent \textbf{Implementation details. }
We collect high-quality gameplay data to fine-tune the Planner, Executor, and Planner's Verifier. Specifically, these data are merged into a single, unified dataset to train an LLM, enabling it to learn both decision-making and self-correction strategies. Details on the SFT process are provided in Appendix \ref{app:method_sft}.




%% file: table/main_results.tex
\begin{table*}[t]
\centering
\caption{Performance of \method without SFT with different models: 
\ding{172} Qwen2.5-7B-Instruct, \ding{173} Qwen3-8B (no\_think), \ding{174} DeepSeek-V3-0324, 
and \ding{172}-SFT: \method with Qwen2.5-7B-Instruct, fine-tuned as Section \ref{subsec:self_improvement}. 
We report the average results of 20 TvT, 20 PvP, and 20 ZvZ games.
$\uparrow$ and $\downarrow$ indicate that higher or lower values represent better performance.
}
\label{tab:main_results}
\begin{tabular}{c|c|cccccc}
\toprule
Built-in AI Level & Model & Win Rate(\%) $\uparrow$ & TCW(s) $\downarrow$ & SBR(\%) $\downarrow$ & RUR(\%) $\uparrow$ & TPD $\downarrow$ & VAR(\%) $\uparrow$ \\
\midrule
\multirow{4}{*}{Harder (LV6)} 
& \ding{172} & 55.00 \small$\pm$ 10.00 & 812 \small$\pm$ 116 & 4.83 \small$\pm$ 3.30 & 7.63 \small$\pm$ 1.36 & 2340 \small$\pm$ 59 & 47.78 \small$\pm$ 3.95 \\
& \ding{173} & 45.00 \small$\pm$ 8.50 & 477 \small$\pm$ 38 & 4.24 \small$\pm$ 2.21 & 8.49 \small$\pm$ 1.17 & 1296 \small$\pm$ 45 & 72.96 \small$\pm$ 3.60 \\
& \ding{174} & 72.88 \small$\pm$ 6.18 & 382 \small$\pm$ 21 & 2.82 \small$\pm$ 0.55 & 11.08 \small$\pm$ 1.01 & 1437 \small$\pm$ 90 & 88.53 \small$\pm$ 3.39 \\
\cmidrule{2-8}
& \ding{172}-SFT & 71.67 \small$\pm$ 10.00 & 410 \small$\pm$ 38 & 4.61 \small$\pm$ 1.22 & 8.62 \small$\pm$ 0.81 & 1398 \small$\pm$ 62 & 85.52 \small$\pm$ 3.74 \\

\midrule
\multirow{4}{*}{VeryHard (LV7)} 
& \ding{172} & 55.00 \small$\pm$ 8.50 & 770 \small$\pm$ 116 & 5.90 \small$\pm$ 2.48 & 9.21 \small$\pm$ 1.07 & 2343 \small$\pm$ 70 & 48.04 \small$\pm$ 2.48 \\
& \ding{173} & 55.00 \small$\pm$ 12.47 & 422 \small$\pm$ 32 & 3.27 \small$\pm$ 1.14 & 9.02 \small$\pm$ 0.44 & 1315 \small$\pm$ 35 & 70.07 \small$\pm$ 4.14 \\
& \ding{174} & 88.33 \small$\pm$ 12.47 & 345 \small$\pm$ 19 & 2.85 \small$\pm$ 1.30 & 11.60 \small$\pm$ 0.84 & 1719 \small$\pm$ 82 & 88.89 \small$\pm$ 3.47 \\
\cmidrule{2-8}
& \ding{172}-SFT & 73.33 \small$\pm$ 14.34 & 419 \small$\pm$ 40 & 5.15 \small$\pm$ 1.15 & 9.54 \small$\pm$ 1.82 & 1427 \small$\pm$ 71 & 84.59 \small$\pm$ 3.72 \\
\bottomrule
\end{tabular}
\end{table*}

%% file: text/5_experiment.tex
\section{Experiments}

In this section, we conduct comprehensive experiments to address three key research questions:  
\textit{Q1}: 
Can \benchmark effectively evaluate the performance of LLM-driven agents in decision-making scenarios?
\textit{Q2}: Does \method consistently generate effective and robust strategies while improving performance through self-improvement mechanisms? 
\textit{Q3}: What are the contributions of each modules in \benchmark and \method to overall performance?

\subsection{Setup}


In this subsection, we outline the experimental setup. 
For the agent-vs-agent evaluation, we employ DeepSeek-V3-0324 \cite{liu2024deepseek}, Qwen2.5-7B-Instruct \cite{qwen2.5}, and Qwen3-8B \cite{qwen3technicalreport} models with simple, predefined prompts as baselines, as details in Appendix \ref{app:simple_agent}. 
For the agent-vs-built-in AI evaluation, we compete against Harder (LV6) and VeryHard (LV7) opponents, which represent the highest non-cheating difficulty levels. 
These results are compared to prior works \cite{ma2024large}. 
Performance is evaluated using metrics defined in Section \ref{subsec:benchmark_metrics}. 
Specifically, Elo ratings are initialized at 1000 for all agents and updated using a K-factor of 32 based on match outcomes. 
Appendix \ref{app:exp_setup} provides further details.

\input{table/verifier_ablation}

\input{table/SFT_score_ablation}

\input{table/comparison_prior}

\subsection{Main Results}
\label{subsec:exp_benchmark}


\vspace{3pt}
\noindent \textbf{Agent-vs-Agent. }
To evaluate the performance of \method against baseline agents, we conduct pairwise matches across 5 games each of Terran vs. Terran (TvT), Zerg vs. Zerg (ZvZ), and Protoss vs. Protoss (PvP), using the outcomes to calculate Elo ratings. 
As shown in Fig. \ref{fig:elo_results}, \method (SFT) achieves the highest Elo rating of 1211 for Terran, outperforming DeepSeek.
For Protoss and Zerg, \method (SFT) also achieves competitive ratings of 1080 and 1096. 
These results highlight three key insights:
\ding{172} \benchmark effectively supports Elo-based agent ranking;
\ding{173} \method demonstrates strong performance, outperforming most baselines;
and \ding{174} fine-tuning significantly enhances performance, enabling \method to match or surpass DeepSeek in specific scenarios.

\vspace{3pt}
\noindent \textbf{Agent-vs-Built-In AI. }
Furthermore, we evaluated the performance of \method without SFT using Qwen2.5-7B-Instruct, Qwen3-8B (no\_think), DeepSeek-V3-0324, and fine-tuned \method with Qwen2.5-7B-Instruct, following the fine-tuning process in Section \ref{subsec:self_improvement}. 
We use metrics in Section \ref{subsec:benchmark_metrics}, excluding Elo, for evaluation. 
As shown in Table \ref{tab:main_results}, \method consistently outperforms challenging built-in AIs across diverse LLMs, demonstrating its robustness. 
Fine-tuning further improves performance: for Qwen2.5-7B-Instruct, fine-tuning increases the win rate from 55\% to 71\% at LV6 and reduces token consumption from 2298 to 1451. 
At the more challenging LV7, the fine-tuned Qwen2.5-7B-Instruct model achieves a 73\% win rate, approaching DeepSeek's performance. 
These findings demonstrate that \method is effective across models and scenarios, with significant improvements after fine-tuning.

\vspace{3pt}
\noindent \textbf{Comparison with prior work. }
We compare our method with \citet{ma2024large} in the Protoss vs. Zerg setting, the primary focus of their study, using 20 matches. 
As shown in Table \ref{tab:comparison_prior}, despite leveraging stronger GPT-based LLMs, \citet{ma2024large} achieves lower win rates than our approach. 
Notably, \citet{ma2024large} simplifies the problem: their agent generates only high-level decisions, while low-level actions are executed via predefined scripts, thereby reducing game complexity. 
In contrast, our method outperforms this baseline with a smaller model while addressing the more challenging full low-level action space.



\subsection{Ablation Study}
\label{subsec:ablation_study}

\noindent \textbf{\benchmark's optimized observation. }
To evaluate the impact of our observation optimization techniques, we conducted ablation experiments by removing two key components: unit ordering and aggregation. 
We performed 20 TvT, ZvZ, and PvP matches for each configuration and analyzed the results. 
As shown in Table \ref{tab:benchmark_ablation} in Appendix \ref{app:additional_experiments}, removing these techniques significantly degrades performance. 
Specifically, removing unit ordering impairs spatial relationship modeling or aggregation, reducing the Valid Action Ratio (VAR). 
This suggests the critical role of both techniques in enabling efficient decision-making within \scii's complex environment.

\vspace{3pt}
\noindent \textbf{Self-correction mechanism and the Planner. }
We ablated the Verifier and Planner modules to assess their contributions by conducting 20 TvT, ZvZ, and PvP matches for each configuration. 
As shown in Table \ref{tab:verifier_ablation}, disabling the Verifier module reduced the win rate by 18\% and VAR by 30\%. 
Removing the Planner module led to a 15\% win rate in all matches. 
This shows the essential role of hierarchical self-correction and strategic planning in robust decision-making.

\vspace{3pt}
\noindent \textbf{Scoring function for SFT data collection. }
To evaluate the effectiveness of the scoring function, we trained on all winning matches instead of using the function to filter experimental data. 
We conducted 20 TvT matches and analyzed the results. 
As shown in Table \ref{tab:tvt_after_SFT}, removing the scoring function significantly decreases performance, with win rates dropping from 73\% to 41\% and VAR decreasing by 33\%. 
These results highlight that, even when distilling high-quality data from high-performance models like DeepSeek, selecting the best data remains essential. 
Our scoring function effectively identifies key game indicators, ensuring better data selection.

\subsection{Visualizations of Gameplay}

In this subsection, we demonstrate \method's strategic planning and tactical execution through gameplay visualizations. 
As shown in Fig. \ref{fig:visualization_terran}(a-c), the Terran agent transitions mid-game advantages into military strength (Fig. \ref{fig:visualization_terran}(b)) and secures victory with a decisive offensive (Fig. \ref{fig:visualization_terran}(c)). 
This illustrates \method's capacity to adapt and strategize effectively in complex environments. 
More visualizations of Protoss and Zerg agents are provided in Appendix \ref{app:visualization}.



%% file: table/verifier_ablation.tex
\begin{table}[t]
\centering
\caption{Performance comparison of \method with Qwen2.5-7B-Instruct, with and without \ding{172} Verifiers and \ding{173} Planner. We report the results of 20 TvT, 20 ZvZ, and 20 PvP games, vs. VeryHard level AI.}
\label{tab:verifier_ablation}
\begin{tabular}{cc|ll}
\toprule
Verifiers & Planner & 
\makecell[c]{Win Rate(\%)} & 
\makecell[c]{VAR(\%)} \\
\midrule
\cmark & \cmark & 55.00 \small$\pm$ 14.53 & 48.04 \small$\pm$ 2.48 \\
\cmark & \xmark & 40.00 \small$\pm$ 8.16 & 50.59 \small$\pm$ 7.67 \\
\midrule
\xmark & \cmark & 36.67 \small$\pm$ 14.53 & 18.62 \small$\pm$ 1.13 \\
\xmark & \xmark & 1.67 \small$\pm$ 3.33 & 20.77 \small$\pm$ 4.42 \\
\bottomrule
\end{tabular}
\end{table}

%% file: table/SFT_score_ablation.tex
\begin{table}[t]
\centering
\caption{Performance comparison of \method with fine-tuned Qwen2.5-7B-Instruct, collecting SFT data with and without scoring function. We evaluate each agent in 20 TvT games, vs. VeryHard AI. }
\label{tab:tvt_after_SFT}
\begin{tabular}{c|ll}
\toprule
Scoring function & 
\makecell[c]{Win Rate(\%)} & 
\makecell[c]{VAR(\%)} \\
\midrule
\cmark & 73.33 \small$\pm$ 14.34 & 84.59 \small$\pm$ 3.72 \\
\xmark & 41.67 \small$\pm$ 8.32 & 51.48 \small$\pm$ 3.55 \\
\bottomrule
\end{tabular}
\end{table}

%% file: table/comparison_prior.tex

\begin{table*}[t]
\centering
\caption{Performance of \method and prior work on Protoss vs. Zerg. We evaluate 20 games, the same as prior work. }
\label{tab:comparison_prior}
\begin{tabular}{cccccc}
\toprule
Method & SFT & Model & \small Harder Win Rate (\%) & \small VeryHard Win Rate (\%) \\
\midrule
\multirow{2}{*}{\method (ours)} 
& \xmark & Qwen2.5-7B-Instruct & 70.00 \small$\pm$ 10.00 & 85.00 \small$\pm$ 20.00 \\
& \cmark & Qwen2.5-7B-Instruct & 85.00 \small$\pm$ 12.25 & 95.00 \small$\pm$ 10.00 \\
\midrule
\multirow{2}{*}{TextStarCraft II~\cite{ma2024large}} 
& \xmark & GPT3.5-Turbo-16K & 50.00 & 8.33 \\
& \cmark & Qwen-7B & 45.00 & - \\
\bottomrule
\end{tabular}
\end{table*}

%% file: text/6_conclusion.tex
\section{Conclusion}

In this paper, we introduce \benchmark, a comprehensive StarCraft II benchmark for evaluating LLMs agents in decision-making tasks. 
\benchmark supports full-length game contexts, diverse race configurations, low-level action spaces, and agent-vs-agent gameplay, enabling robust performance assessment and ranking. 
We also present \method, a closed-loop agent framework that combines hierarchical planning with iterative self-correction and continuous self-improvement through supervised fine-tuning. 
Experiments demonstrate the effectiveness of \benchmark for agent evaluation and highlight \method's ability to achieve strong performance, including enabling smaller models to compete effectively against challenging opponents. 
These results show the potential of \benchmark and \method to advance research in complex decision-making for LLMs.

%% file: appendix/benchmark_detail.tex
\section{Implementation of \benchmark}

\subsection{Introduction to \scii}
\label{app:scii_intro}



In this subsection, we provide a concise overview of StarCraft II, a landmark real-time strategy (RTS) game developed by Blizzard Entertainment. Renowned for its strategic complexity, tactical depth, and asymmetrical balance between three distinct races, StarCraft II has become a benchmark for AI research in dynamic, partially observable environments requiring real-time decision-making.

\paragraph{Basic Interface. } 
The game features a top-down view of the battlefield, with a primary map display, a minimap for global awareness, and mouse-and-keyboard controls. Players use the mouse to select units, issue commands, and navigate the map, while the keyboard enables rapid actions through hotkeys, such as unit production and camera movement. The minimap allows monitoring of multiple locations simultaneously, facilitating quick responses. Resources, including minerals and vespene gas, are displayed alongside a command panel that provides access to unit and building functions.

\paragraph{Game Objectives and Core Gameplay. }
The objective is to defeat opponents by destroying their bases while protecting one's own. Success requires a combination of strategic planning and tactical execution:
\begin{itemize}
\item \textbf{Strategy: }
Players manage resource collection, base expansion, and technological progression to unlock advanced units and upgrades. Securing additional resource nodes is essential for sustained production, while efficient resource allocation ensures competitiveness over time.
\item \textbf{Tactics: }
Combat relies on precise unit control, such as positioning, micro-management (e.g., dodging attacks or optimizing unit abilities), and multi-tasking across multiple fronts. The fog of war, which obscures unscouted areas, adds to the challenge by limiting available information, requiring adaptive decision-making under time constraints.
\end{itemize}

\paragraph{Race Mechanisms. }
The game's three asymmetric races—Terran, Zerg, and Protoss—offer distinct playstyles while maintaining competitive balance:
\begin{itemize}
\item \textbf{Terran: } 
A versatile, mechanically focused race with strong defensive capabilities. Terran units and buildings can be repaired, and structures like bunkers and siege tanks excel at controlling key areas. Add-ons and mobile buildings enhance adaptability, enabling positional advantages.
\item \textbf{Zerg: } 
An aggressive, swarm-based race that emphasizes rapid production and map control. Units are spawned from larvae at hatcheries, requiring efficient resource management and expansion. Zerg structures must be built on creep, an organic terrain that spreads dynamically, encouraging offensive gameplay.
\item \textbf{Protoss: } 
A technologically advanced race featuring powerful but costly units. Protoss structures require energy supplied by pylons, and units are warped in rather than produced conventionally. Shields that regenerate over time promote hit-and-run tactics, while warp gates allow rapid reinforcement, favoring decisive engagements.
\end{itemize}

The combination of economic strategy, technological progression, and tactical combat, along with the unique characteristics of each race, creates a highly dynamic and competitive environment. StarCraft II epitomizes the challenges of real-time multi-agent systems, making it an enduring platform for AI research.

\subsection{Formulation of \benchmark's Observation and Action}
\label{app:obs_example}

\benchmark provides a standardized text-based interface designed for LLM agents. Observations are presented as hierarchical textual summaries of the game state, employing two key optimizations to minimize cognitive load while retaining essential information: 
\begin{itemize}
\item \textbf{Unit aggregation}, which groups units performing similar tasks (e.g., mining SCVs), and 
\item \textbf{Proximity-based unit ordering}, inspired by the Traveling Salesman Problem (TSP), which sequences units and buildings based on spatial proximity, starting from the Command Center. This structure forms an explicit topological map, aiding LLMs in understanding functional dependencies (e.g., Barracks-TechLab) and battlefield positions without processing raw game data. Actions are encoded in a unified JSON format, allowing for precise multi-unit commands.
\end{itemize}


\paragraph{Observation Example. }
The textualized observation encompasses the following information: current game settings (time, race, map), resource status (minerals, vespene, supply), the state of allied units, abilities of allied units/buildings, the status of visible enemy units/buildings, descriptions of allied abilities, history of allied actions, and the map resources distribution. 
To facilitate model decisions with supplementary information, \benchmark incorporates brief descriptions of the tech tree and available actions specific. 
An example is provided as follows:

\begin{promptbox}{Textual Observation Example}
\# Round state\\
Time: 04:18\\
Race: Protoss\\
Minerals: 175\\
Vespene: 154\\
Supply army: 18\\
Supply workers: 18\\
Supply unused: 3\\
Map size: 48x48\\

\# Own units\\
\lbrack 9, 809, 953, 569, 665, 857, 721, 529, 281, 385, 818, 145, 105, 129, 521, 697, 97\rbrack Probe\\
State: collecting resources automatically\\
\lbrack 399\rbrack Stalker\\
Position: (43, 17)\\
Health: 80/80 (100\%)\\
Shield: 80/80\\
State: idle\\
\lbrack 539\rbrack Zealot\\
Position: (40, 13)\\
Health: 100/100 (100\%)\\
Shield: 50/50\\
State: idle\\
\lbrack 434\rbrack Stalker\\
Position: (44, 36)\\
Health: 80/80 (100\%)\\
Shield: 80/80\\
State: attacking \lbrack 249\rbrack Nexus\\
\lbrack 260\rbrack Zealot\\
Position: (50, 38)\\
Health: 100/100 (100\%)\\
Shield: 50/50\\
State: attacking \lbrack 705\rbrack Gateway\\
\lbrack 986\rbrack Zealot\\
Position: (46, 42)\\
Health: 100/100 (100\%)\\
Shield: 50/50\\
State: attacking \lbrack 249\rbrack Nexus\\
\lbrack 122\rbrack Zealot\\
Position: (49, 47)\\
Health: 100/100 (100\%)\\
Shield: 50/50\\
State: attacking \lbrack 249\rbrack Nexus\\
\lbrack 2\rbrack Zealot\\
Position: (48, 48)\\
Health: 100/100 (100\%)\\
Shield: 50/50\\
State: attacking \lbrack 249\rbrack Nexus\\

\# Unit abilities\\
Probe\lbrack 9, 809, 953, 569, 665, 857, 721, 529, 281, 385, 818, 145, 105, 129, 521, 697, 97\rbrack : PROTOSSBUILD\_NEXUS, PROTOSSBUILD\_PYLON, PROTOSSBUILD\_ASSIMILATOR, PROTOSSBUILD\_GATEWAY, PROTOSSBUILD\_FORGE, PROTOSSBUILD\_TWILIGHTCOUNCIL, PROTOSSBUILD\_STARGATE, PROTOSSBUILD\_ROBOTICSFACILITY, PROTOSSBUILD\_CYBERNETICSCORE, BUILD\_SHIELDBATTERY\\
Zealot\lbrack 986, 539, 260, 122, 2\rbrack : MOVE\_MOVE, ATTACK\_ATTACK\\
Stalker\lbrack 399, 434\rbrack : MOVE\_MOVE, ATTACK\_ATTACK\\

\# Own structures\\
\lbrack 377\rbrack Nexus\\
Position: (47, 22)\\
Health: 1000/1000 (100\%)\\
Shield: 1000/1000\\
Energy: 53/200\\
State: idle\\
Harvesters: 15/16\\
\lbrack 265\rbrack Pylon\\
Position: (48, 19)\\
Health: 200/200 (100\%)\\
Shield: 200/200\\
State: idle\\
\lbrack 273\rbrack Gateway\\
Position: (43, 19)\\
Health: 500/500 (100\%)\\
Shield: 500/500\\
Production list: Stalker, Stalker, Stalker, Stalker\\
\lbrack 713\rbrack Pylon\\
Position: (41, 19)\\
Health: 136/200 (68\%)\\
Shield: 200/200\\
State: idle\\
\lbrack 554\rbrack Gateway\\
Position: (40, 15)\\
Health: 500/500 (100\%)\\
Shield: 500/500\\
Production list: Zealot, Zealot\\
\lbrack 978\rbrack Pylon\\
Position: (42, 13)\\
Health: 200/200 (100\%)\\
Shield: 200/200\\
State: idle\\
\lbrack 289\rbrack CyberneticsCore\\
Position: (46, 13)\\
Health: 550/550 (100\%)\\
Shield: 550/550\\
State: idle\\
\lbrack 841\rbrack Assimilator\\
Position: (54, 25)\\
Health: 300/300 (100\%)\\
Shield: 300/300\\
State: idle\\
Harvesters: 3/3 (no more harvesters accepted)\\

\# Structure abilities\\
CyberneticsCore\lbrack 289\rbrack : CYBERNETICSCORERESEARCH\_PROTOSSAIRWEAPONSLEVEL1, CYBERNETICSCORERESEARCH\_PROTOSSAIRARMORLEVEL1\\
Nexus\lbrack 377\rbrack : NEXUSTRAIN\_PROBE, EFFECT\_CHRONOBOOSTENERGYCOST, EFFECT\_MASSRECALL\_NEXUS\\
Gateway\lbrack 273, 554\rbrack : GATEWAYTRAIN\_ZEALOT, GATEWAYTRAIN\_STALKER, GATEWAYTRAIN\_SENTRY, TRAIN\_ADEPT\\

\# Visible enemy units\\
\lbrack Empty\rbrack 

\# Visible enemy structures\\
\lbrack 705\rbrack Gateway\\
Position: (50, 40)\\
Health: 500/500 (100\%)\\
Shield: 487/500\\
\lbrack 561\rbrack Assimilator\\
Position: (54, 42)\\
Health: 300/300 (100\%)\\
Shield: 300/300\\
\lbrack 249\rbrack Nexus\\
Position: (47, 45)\\
Health: 1000/1000 (100\%)\\
Shield: 136/1000\\
Energy: 53/200\\
\lbrack 429\rbrack Pylon\\
Position: (37, 50)\\

\# Action history
\begin{verbatim}
{"action": "GATEWAYTRAIN_STALKER", "units": [273]}
{"action": "GATEWAYTRAIN_ZEALOT", "units": [554]}
{"action": "ATTACK_ATTACK", "units": [434], "target_unit": 418}
{"action": "ATTACK_ATTACK", "units": [986], "target_unit": 242}
{"action": "GATEWAYTRAIN_STALKER", "units": [273]}
{"action": "MOVE_MOVE", "units": [2], "target_position": [47, 45]}
{"action": "GATEWAYTRAIN_STALKER", "units": [273]}
{"action": "ATTACK_ATTACK", "units": [434], "target_unit": 249}
{"action": "ATTACK_ATTACK", "units": [986], "target_unit": 249}
{"action": "GATEWAYTRAIN_STALKER", "units": [273]}
\end{verbatim}

\# Map information\\
No mineral fields found\\
Closest vespene geysers: [321](54, 25), [729](43, 15), [193](54, 42), [625](43, 52)\\

\# Ability description\\
ATTACK\_ATTACK(target: PointOrUnit): Attack some unit or structure.\\
MOVE\_MOVE(target: PointOrUnit): Move to target position.\\
MOVE(target: PointOrUnit): Move to target position.\\
PROTOSSBUILD\_ASSIMILATOR(target: Unit): Build an assimilator, built on a Vespene Geyser that allows Probes to harvest gas. Cost: 75 minerals.\\
PROTOSSBUILD\_CYBERNETICSCORE(target: Point): Build a Cybernetics Core. Cost: 150 minerals.\\
PROTOSSBUILD\_FORGE(target: Point): Build a Forge. Cost: 150 minerals.\\
PROTOSSBUILD\_GATEWAY(target: Point): Build a Gateway to training ground units. Cost: 150 minerals.\\
PROTOSSBUILD\_NEXUS(target: Point): Build a Nexus. Cost: 400 minerals.\\
PROTOSSBUILD\_PYLON(target: Point): Build a Pylon, providing supply and projects a power field. Cost: 100 minerals.\\
PROTOSSBUILD\_ROBOTICSFACILITY(target: Point): Build a Robotics Facility. Cost: 150 minerals, 100 vespene.\\
PROTOSSBUILD\_STARGATE(target: Point): Build a Protoss Stargate. Cost: 150 minerals, 150 vespene.\\
PROTOSSBUILD\_TWILIGHTCOUNCIL(target: Point): Build a Twilight Council. Cost: 150 minerals, 100 vespene.\\
BUILD\_SHIELDBATTERY(target: Point): Build a Shield Battery to restore shields of a single friendly unit or building within 6 range. Cost: 100 minerals.\\
NEXUSTRAIN\_PROBE(target: None): Train a Probe. Cost: 50 minerals.\\
EFFECT\_CHRONOBOOSTENERGYCOST(target: Unit): Reduce the unit production time or technological research time of the target building.\\
EFFECT\_CHRONOBOOST(target: None): Reduce the unit production time or technological research time of the target building.\\
EFFECT\_MASSRECALL(target: Point): Instantly teleport the unit back to its current location.\\
EFFECT\_MASSRECALL\_NEXUS(target: Point): Instantly teleport the unit back to its current location.\\
GATEWAYTRAIN\_ZEALOT(target: None): Train a Zealot, a durable frontline warrior that charges to rapidly close with and attack ground enemies. Cost: 100 minerals.\\
GATEWAYTRAIN\_STALKER(target: None): Train a Stalker, a mobile ranged warrior that can Blink to outmaneuver foes and attack both ground and air targets. Cost: 125 minerals, 50 vespene.\\
GATEWAYTRAIN\_SENTRY(target: None): Train a Sentry, a tactical support caster that manipulates the battlefield with Force Fields and protects allies with a Guardian Shield. Cost: 50 minerals, 100 vespene.\\
TRAIN\_ADEPT(target: None): Train an Adept, a ranged warrior that projects a psionic shade to teleport past enemy forces and harass light units. Cost: 100 minerals, 25 vespene.\\
CYBERNETICSCORERESEARCH\_PROTOSSAIRWEAPONSLEVEL1(target: None): Upgrades the damage of Protoss air units. Cost: 100 minerals, 100 vespene.\\
CYBERNETICSCORERESEARCH\_PROTOSSAIRARMORLEVEL1(target: None): Upgrades the armor of Protoss air units. Cost: 150 minerals, 150 vespene.\\
\end{promptbox}
\label{fig:obs_example}

\paragraph{Action Example. }
In \benchmark, executable actions are represented as a list of JSON objects. Every action object requires two primary keys: \texttt{action}, specifying the command, and \texttt{units}, indicating the unit(s) involved. For specific actions, additional keys such as \texttt{target\_unit} or \texttt{target\_position} are needed to specify the target entity or location. An example is provided asffollows:

\begin{promptbox}{Textual Action Example}

\begin{verbatim}
[
    {
        "action": "ATTACK_ATTACK",
        "units": [1, 2, 3],
        "target_unit": 9
    },
    {
        "action": "MOVE_MOVE",
        "units": [4, 5],
        "target_position": [50, 60]
    },
    {
        "action": "COMMANDCENTERTRAIN_SCV",
        "units": [6]
    }
]
\end{verbatim}
\end{promptbox}
\label{fig:action_example}

\subsection{\benchmark's Implementation Details}
\label{app:benchmark_implementation}

\paragraph{How \benchmark Interacts with StarCraft II. }
\benchmark is built on the widely used Python-sc2 library, which provides a Python API for interacting with the StarCraft II game client via its raw API. 
We extended this library to create a standardized environment for LLM-based agents by enabling real-time conversion of complex game states into structured text observations, and translating the LLM's JSON-formatted actions into executable game commands. 

\paragraph{Computational Resources Required. }
Running experiments in \benchmark involves two key components: the agent (LLM) and the environment (game client).
\textbf{Agent (LLM Server)}:  Our experiments used a server with an AMD EPYC 64-Core CPU, 503GB RAM, and six NVIDIA A100 (80GB) GPUs, running Ubuntu 18.04, Python 3.9+, and necessary deep learning frameworks. Detailed environmental requirements are provided in our anonymous code repository.  
\textbf{Environment (Game Client)}: The StarCraft II game client can run on a standard desktop or laptop computer. A local installation of StarCraft II on a compatible operating system (Windows or macOS) is required.

\subsection{\benchmark's Performance Metrics}
\label{app:metrics}

In this subsection, we describe the evaluation metrics used for \benchmark. 
First, we explain the calculation of the Elo rating.
Then, we introduce the other metrics and provide their formulation.

\paragraph{Elo Rating System. }
We employ the standard Elo rating system to quantify agent performance in agent-vs-agent comparisons. Each agent starts with an initial rating of $R_0 = 1000$. After a match between agent $A$ (rating $R_A$) and agent $B$ (rating $R_B$), their ratings update as:
\begin{enumerate}
    \item \textbf{Expected Score Calculation:}
    $A$'s expected score is derived from the rating difference:
    \[
    E_A = \frac{1}{1 + 10^{(R_B - R_A)/400}}
    \]
    $B$'s expected score is $E_B = 1 - E_A$.
    
    \item \textbf{Actual Score Assignment:}
    The outcome $S_A$ is assigned as:
    \begin{equation*}
    \begin{aligned}
    S_A & = \begin{cases} 
      1.0 & \text{win} \\
      0.5 & \text{tie} \\
      0.0 & \text{loss}
    \end{cases},
    \\
    S_B & = 1 - S_A.
    \end{aligned}
    \end{equation*}
    
    \item \textbf{Rating Update:}
    New ratings are calculated using:
    \begin{equation*}
    \begin{aligned}
    R'_A & = R_A + K \cdot (S_A - E_A),
    \\
    R'_B & = R_B + K \cdot (S_B - E_B),
    \end{aligned}
    \end{equation*}
    where $K=32$ is the update sensitivity factor.
\end{enumerate}
Matches are processed in random order. Ratings update sequentially after each match, with new agents initialized to $R_0$ upon first appearance. Final rankings are determined by descending order of converged ratings.

\paragraph{Other metrics. }
Then, we introduce the other metrics and provide their formulation.

\begin{enumerate}
  \item \textbf{Win Rate (WR)}: The proportion of games won against a fixed set of opponents (e.g., built-in AIs). It reflects overall effectiveness in achieving victory under standardized conditions.
  \[
    \mathrm{WR} = \frac{\text{Number of Wins}}{\text{Total Games Played}} \times 100\%
  \]

  \item \textbf{Time Cost of Winning (TCW)}: Average duration of victorious games, measured in game seconds. It indicates strategic efficiency; lower TCW implies faster execution of winning strategies.
  \[
    \mathrm{TCW} = \frac{1}{N_{\mathrm{wins}}} \sum_{i=1}^{N_{\mathrm{wins}}} T_i
  \]
  where $T_i$ is the duration of the $i$-th winning game.

  \item \textbf{Supply Block Ratio (SBR)}: Fraction of game time during which unit production is halted due to insufficient supply capacity. It evaluates macro-management skill; lower SBR denotes optimal resource allocation and production continuity.
  \[
    \mathrm{SBR} = \frac{\text{Time Supply Capped}}{\text{Total Game Time}} \times 100\%
  \]

  \item \textbf{Resource Utilization Ratio (RUR)}: Average resources (minerals + gas) spent per game timestep. It quantifies economic efficiency; higher RUR indicates effective resource conversion into units/upgrades.
  \[
    \mathrm{RUR} = \frac{\text{Total Resources Spent}}{\text{Total Game Timesteps}}
  \]

  \item \textbf{Tokens Per Decision (TPD)}: Average output token count consumed by the LLM to generate a single in-game decision. It measures computational efficiency of the LLM agent; lower TPD implies leaner prompt engineering and faster inference.
  \[
    \mathrm{TPD} = \frac{\text{Total Tokens Processed}}{\text{Total Decisions Made}}
  \]

  \item \textbf{Valid Action Ratio (VAR)}: Proportion of agent decisions that are syntactically and semantically executable in the game environment. It assesses action robustness; higher VAR indicates better alignment with game rules and context awareness.
  \[
    \mathrm{VAR} = \frac{\text{Valid Decisions}}{\text{Total Decisions Attempted}} \times 100\%
  \]
\end{enumerate}


%% file: appendix/method_detail.tex
\vspace{10pt}
\section{Implementation of \method}

\subsection{Planner}
\label{app:planner_prompt}

The \method Planner analyzes the current game state and generates natural language command directives using a structured prompt template. In this template, \texttt{obs\_text} represents the textualized game state, \texttt{rules\_list} is a dynamic list of expert knowledge, and \texttt{plan\_example} consists of plan examples for different races.

\begin{promptbox}{Prompt Template of Planner}
As a top-tier StarCraft II strategist, your task is to give one or more commands based on the current game state. Only give commands which can be executed immediately, instead of waiting for certain events.\\

\textbf{Aim}\\
Our final aim: destroy all enemies as soon as possible.\\
Our strategy:\\
- Resource collection: produce workers and gather minerals and gas\\
- Development: build attacking units and structures\\
- Attacking: concentrate forces to search and destroy enemies proactively\\

\textbf{Current Game State}\\
\{obs\_text\}\\

\textbf{Rules}\\
\{rules\_list\}\\

\textbf{Examples}\\
\{plan\_example\}\\

Think step by step, and then give commands as a list JSON in the following format wrapped with triple backticks:
\begin{verbatim}
```
[
    "<command_1>",
    "<command_2>",
    ...
]
```
\end{verbatim}
\end{promptbox}

Below is a complete prompt example from a Protoss vs. Protoss match.

\begin{promptbox}{Prompt Example of Planner}
As a top-tier StarCraft II strategist, your task is to give one or more commands based on the current game state. Only give commands which can be executed immediately, instead of waiting for certain events.\\

\textbf{Aim}\\
Our final aim: destroy all enemies as soon as possible.\\
Our strategy:\\
- Resource collection: produce workers and gather minerals and gas\\
- Development: build attacking units and structures\\
- Attacking: concentrate forces to search and destroy enemies proactively\\

\textbf{Current Game State}\\
\# Round state\\
Time: 04:18\\
Race: Protoss\\
Minerals: 175\\
Vespene: 154\\
Supply army: 18\\
Supply workers: 18\\
Supply unused: 3\\
Map size: 48x48\\

\# Own units\\
\lbrack 9, 809, 953, 569, 665, 857, 721, 529, 281, 385, 818, 145, 105, 129, 521, 697, 97\rbrack Probe\\
State: collecting resources automatically\\
\lbrack 399\rbrack Stalker\\
Position: (43, 17)\\
Health: 80/80 (100\%)\\
Shield: 80/80\\
State: idle\\
\lbrack 539\rbrack Zealot\\
Position: (40, 13)\\
Health: 100/100 (100\%)\\
Shield: 50/50\\
State: idle\\
\lbrack 434\rbrack Stalker\\
Position: (44, 36)\\
Health: 80/80 (100\%)\\
Shield: 80/80\\
State: attacking \lbrack 249\rbrack Nexus\\
\lbrack 260\rbrack Zealot\\
Position: (50, 38)\\
Health: 100/100 (100\%)\\
Shield: 50/50\\
State: attacking \lbrack 705\rbrack Gateway\\
\lbrack 986\rbrack Zealot\\
Position: (46, 42)\\
Health: 100/100 (100\%)\\
Shield: 50/50\\
State: attacking \lbrack 249\rbrack Nexus\\
\lbrack 122\rbrack Zealot\\
Position: (49, 47)\\
Health: 100/100 (100\%)\\
Shield: 50/50\\
State: attacking \lbrack 249\rbrack Nexus\\
\lbrack 2\rbrack Zealot\\
Position: (48, 48)\\
Health: 100/100 (100\%)\\
Shield: 50/50\\
State: attacking \lbrack 249\rbrack Nexus\\

\# Unit abilities\\
Probe\lbrack 9, 809, 953, 569, 665, 857, 721, 529, 281, 385, 818, 145, 105, 129, 521, 697, 97\rbrack : PROTOSSBUILD\_NEXUS, PROTOSSBUILD\_PYLON, PROTOSSBUILD\_ASSIMILATOR, PROTOSSBUILD\_GATEWAY, PROTOSSBUILD\_FORGE, PROTOSSBUILD\_TWILIGHTCOUNCIL, PROTOSSBUILD\_STARGATE, PROTOSSBUILD\_ROBOTICSFACILITY, PROTOSSBUILD\_CYBERNETICSCORE, BUILD\_SHIELDBATTERY\\
Zealot\lbrack 986, 539, 260, 122, 2\rbrack : MOVE\_MOVE, ATTACK\_ATTACK\\
Stalker\lbrack 399, 434\rbrack : MOVE\_MOVE, ATTACK\_ATTACK\\

\# Own structures\\
\lbrack 377\rbrack Nexus\\
Position: (47, 22)\\
Health: 1000/1000 (100\%)\\
Shield: 1000/1000\\
Energy: 53/200\\
State: idle\\
Harvesters: 15/16\\
\lbrack 265\rbrack Pylon\\
Position: (48, 19)\\
Health: 200/200 (100\%)\\
Shield: 200/200\\
State: idle\\
\lbrack 273\rbrack Gateway\\
Position: (43, 19)\\
Health: 500/500 (100\%)\\
Shield: 500/500\\
Production list: Stalker, Stalker, Stalker, Stalker\\
\lbrack 713\rbrack Pylon\\
Position: (41, 19)\\
Health: 136/200 (68\%)\\
Shield: 200/200\\
State: idle\\
\lbrack 554\rbrack Gateway\\
Position: (40, 15)\\
Health: 500/500 (100\%)\\
Shield: 500/500\\
Production list: Zealot, Zealot\\
\lbrack 978\rbrack Pylon\\
Position: (42, 13)\\
Health: 200/200 (100\%)\\
Shield: 200/200\\
State: idle\\
\lbrack 289\rbrack CyberneticsCore\\
Position: (46, 13)\\
Health: 550/550 (100\%)\\
Shield: 550/550\\
State: idle\\
\lbrack 841\rbrack Assimilator\\
Position: (54, 25)\\
Health: 300/300 (100\%)\\
Shield: 300/300\\
State: idle\\
Harvesters: 3/3 (no more harvesters accepted)\\

\# Structure abilities\\
CyberneticsCore\lbrack 289\rbrack : CYBERNETICSCORERESEARCH\_PROTOSSAIRWEAPONSLEVEL1, CYBERNETICSCORERESEARCH\_PROTOSSAIRARMORLEVEL1\\
Nexus\lbrack 377\rbrack : NEXUSTRAIN\_PROBE, EFFECT\_CHRONOBOOSTENERGYCOST, EFFECT\_MASSRECALL\_NEXUS\\
Gateway\lbrack 273, 554\rbrack : GATEWAYTRAIN\_ZEALOT, GATEWAYTRAIN\_STALKER, GATEWAYTRAIN\_SENTRY, TRAIN\_ADEPT\\

\# Visible enemy units\\
\lbrack Empty\rbrack 

\# Visible enemy structures\\
\lbrack 705\rbrack Gateway\\
Position: (50, 40)\\
Health: 500/500 (100\%)\\
Shield: 487/500\\
\lbrack 561\rbrack Assimilator\\
Position: (54, 42)\\
Health: 300/300 (100\%)\\
Shield: 300/300\\
\lbrack 249\rbrack Nexus\\
Position: (47, 45)\\
Health: 1000/1000 (100\%)\\
Shield: 136/1000\\
Energy: 53/200\\
\lbrack 429\rbrack Pylon\\
Position: (37, 50)\\

\# Action history
\begin{verbatim}
{"action": "GATEWAYTRAIN_STALKER", "units": [273]}
{"action": "GATEWAYTRAIN_ZEALOT", "units": [554]}
{"action": "ATTACK_ATTACK", "units": [434], "target_unit": 418}
{"action": "ATTACK_ATTACK", "units": [986], "target_unit": 242}
{"action": "GATEWAYTRAIN_STALKER", "units": [273]}
{"action": "MOVE_MOVE", "units": [2], "target_position": [47, 45]}
{"action": "GATEWAYTRAIN_STALKER", "units": [273]}
{"action": "ATTACK_ATTACK", "units": [434], "target_unit": 249}
{"action": "ATTACK_ATTACK", "units": [986], "target_unit": 249}
{"action": "GATEWAYTRAIN_STALKER", "units": [273]}
\end{verbatim}

\# Map information\\
No mineral fields found\\
Closest vespene geysers: [321](54, 25), [729](43, 15), [193](54, 42), [625](43, 52)\\

\# Ability description\\
ATTACK\_ATTACK(target: PointOrUnit): Attack some unit or structure.\\
MOVE\_MOVE(target: PointOrUnit): Move to target position.\\
MOVE(target: PointOrUnit): Move to target position.\\
PROTOSSBUILD\_ASSIMILATOR(target: Unit): Build an assimilator, built on a Vespene Geyser that allows Probes to harvest gas. Cost: 75 minerals.\\
PROTOSSBUILD\_CYBERNETICSCORE(target: Point): Build a Cybernetics Core. Cost: 150 minerals.\\
PROTOSSBUILD\_FORGE(target: Point): Build a Forge. Cost: 150 minerals.\\
PROTOSSBUILD\_GATEWAY(target: Point): Build a Gateway to training ground units. Cost: 150 minerals.\\
PROTOSSBUILD\_NEXUS(target: Point): Build a Nexus. Cost: 400 minerals.\\
PROTOSSBUILD\_PYLON(target: Point): Build a Pylon, providing supply and projects a power field. Cost: 100 minerals.\\
PROTOSSBUILD\_ROBOTICSFACILITY(target: Point): Build a Robotics Facility. Cost: 150 minerals, 100 vespene.\\
PROTOSSBUILD\_STARGATE(target: Point): Build a Protoss Stargate. Cost: 150 minerals, 150 vespene.\\
PROTOSSBUILD\_TWILIGHTCOUNCIL(target: Point): Build a Twilight Council. Cost: 150 minerals, 100 vespene.\\
BUILD\_SHIELDBATTERY(target: Point): Build a Shield Battery to restore shields of a single friendly unit or building within 6 range. Cost: 100 minerals.\\
NEXUSTRAIN\_PROBE(target: None): Train a Probe. Cost: 50 minerals.\\
EFFECT\_CHRONOBOOSTENERGYCOST(target: Unit): Reduce the unit production time or technological research time of the target building.\\
EFFECT\_CHRONOBOOST(target: None): Reduce the unit production time or technological research time of the target building.\\
EFFECT\_MASSRECALL(target: Point): Instantly teleport the unit back to its current location.\\
EFFECT\_MASSRECALL\_NEXUS(target: Point): Instantly teleport the unit back to its current location.\\
GATEWAYTRAIN\_ZEALOT(target: None): Train a Zealot, a durable frontline warrior that charges to rapidly close with and attack ground enemies. Cost: 100 minerals.\\
GATEWAYTRAIN\_STALKER(target: None): Train a Stalker, a mobile ranged warrior that can Blink to outmaneuver foes and attack both ground and air targets. Cost: 125 minerals, 50 vespene.\\
GATEWAYTRAIN\_SENTRY(target: None): Train a Sentry, a tactical support caster that manipulates the battlefield with Force Fields and protects allies with a Guardian Shield. Cost: 50 minerals, 100 vespene.\\
TRAIN\_ADEPT(target: None): Train an Adept, a ranged warrior that projects a psionic shade to teleport past enemy forces and harass light units. Cost: 100 minerals, 25 vespene.\\
CYBERNETICSCORERESEARCH\_PROTOSSAIRWEAPONSLEVEL1(target: None): Upgrades the damage of Protoss air units. Cost: 100 minerals, 100 vespene.\\
CYBERNETICSCORERESEARCH\_PROTOSSAIRARMORLEVEL1(target: None): Upgrades the armor of Protoss air units. Cost: 150 minerals, 150 vespene.\\

\textbf{Rules}\\
1. Commands should be natural language, instead of code.\\
2. Produce as many units with the strongest attack power as possible.\\
3. The total cost of all commands should not exceed the current resources (minerals and gas).\\
4. Commands should not build redundant structures(e.g. 2 Refinery while one is not fully utilized).\\
5. Commands should not use abilities that are not supported currently.\\
6. Commands should not build a structure that is not needed now (e.g. build a Missile Turret but there is no enemy air unit).\\
7. The unit production list capacity of structures is 5. If the list is full, do not add more units to it.\\
8. Commands should not send Probe to gather resources because the system will do it automatically.\\
9. Commands should not train too many Probes, whose number should not exceed the capacity of Nexus and Assimilator.\\
10. Commands can construct a new one Pylon only when the remaining unused supply is less than 7.\\
11. The enemy will start a fierce attack at 03:00, so you need to start producing a large number of attack units, such as Marauder, at least at 02:30.\\
12. Supply is low! Build a Pylon immediately.\\
13. Some of your structures are unpowered! Build a Pylon nearby.\\
14. Your Nexus has enough energy for Chrono Boost. Use it on the Nexus for more Probes or on a production building.\\
15. Consider building more Gateways to increase unit production.\\
16. Your army has many Zealots. Produce more Stalkers for ranged support.\\

\textbf{Examples}\\
Following are some examples:\\
- Do nothing and just wait;\\
- Train 1/2/3/... Probe/Stalker/Zealot/...\\
- Build a Pylon;\\
- Upgrade to Warp Gate;\\
- Attack visible enemies;\\
- ...\\

Think step by step, and then give commands as a list JSON in the following format wrapped with triple backticks:
\begin{verbatim}
```
[
    "<command_1>",
    "<command_2>",
    ...
]
```
\end{verbatim}
\end{promptbox}

We provide an example output of the Planner from DeepSeek-V3-0324 as follows. It employs a step-by-step reasoning process and outputs five plans in JSON format.

\begin{promptbox}{Planner Output Example by DeepSeek-V3-0324}
Let me analyze the current game state and determine the best immediate actions:\\

1. Supply Situation:\\
- Current supply: 18 (workers) + 18 (army) = 36 used\\
- Total supply: 36 + 3 unused = 39\\
- We have 3 unused supply left (Rule 12 says build Pylon when <7, so not urgent yet)\\

2. Resources:\\
- 175 minerals, 154 vespene\\
- Enough for Stalkers (125/50) or Zealots (100/0)\\

3. Production:\\
- Gateway \lbrack 273\rbrack  has 4 Stalkers in queue (can add 1 more)\\
- Gateway \lbrack 554\rbrack  has 2 Zealots in queue (can add 3 more)\\
- Nexus has 15/16 harvesters (could train 1 more Probe)\\

4. Army:\\
- Attacking enemy Nexus and Gateway with 4 Zealots and 1 Stalker\\
- 1 idle Stalker and 1 idle Zealot that should join the attack\\

5. Structures:\\
- Pylon \lbrack 713\rbrack  is damaged (68\% health) - may need attention soon\\
- All structures appear powered\\
- No immediate need for new production buildings yet\\

6. Chrono Boost:\\
- Nexus has 53 energy (enough for 1 Chrono Boost)\\

Best immediate actions:\\
1. Use Chrono Boost on Gateway \lbrack 273\rbrack  to speed up Stalker production\\
2. Add 1 more Stalker to Gateway \lbrack 273\rbrack  queue (we have resources)\\
3. Add 1 more Zealot to Gateway \lbrack 554\rbrack  queue\\
4. Send idle units to join the attack on enemy Nexus\\

\begin{verbatim}
```
[
    "Use Chrono Boost on Gateway [273]",
    "Train 1 Stalker at Gateway [273]",
    "Train 1 Zealot at Gateway [554]",
    "Attack enemy Nexus [249] with Stalker [399]",
    "Attack enemy Nexus [249] with Zealot [539]"
]
```
\end{verbatim}
\end{promptbox}



\subsection{Planner Verifier}

The Planner's Verifier in \method validates the plans generated by the Planner and identifies errors using a set of dynamic expert knowledge. 
This process employs the prompt template as follows, where \texttt{given\_command} refers to the plans produced by the Planner, and \texttt{rules\_list} is the same as the part of the Planner prompt.

\begin{promptbox}{Prompt Template of Planner Verifier}
As a top-tier StarCraft II player, your task is to check if the given commands for current game state violate any rules.\\

\textbf{Current Game State}\\
\{obs\_text\}\\

\textbf{Given Commands}\\
\{given\_command\}\\

\textbf{Rules Checklist}\\
\{rule\_list\}\\

Analyze the given rules one by one, and then provide a summary for errors at the end as follows, wrapped with triple backticks:\\
\begin{verbatim}
```
{
    "errors": [
        "Error 1: ...",
        "Error 2: ...",
        ...
    ],
    "error_number": 0/1/2/...
}
```
\end{verbatim}
\end{promptbox}

The Verifier operates using a dynamic knowledge list comprising expert rules specific to three races. This list includes a small, static set that is always active and a larger, conditional set triggered by specific in-game events. For instance, the first three shared knowledge entries are permanently included in decision-making, while the 12th Terran entry is added only when supply is low. The full knowledge bases for each race are detailed below.

\begin{promptbox}{Complete Rule Set of Planner Verifier for Terran}
1. Commands should be natural language, instead of code.\\
2. Produce as many units with the strongest attack power as possible.\\
3. The total cost of all commands should not exceed the current resources (minerals and gas).\\
4. Commands should not build redundant structures while the existing ones are idle.\\
5. Commands should not build redundant structures(e.g. 2 Refinery while one is not fully utilized).\\
6. Commands should not use abilities that are not supported currently.\\
7. Commands should not build a structure that is not needed now (e.g. build a Missile Turret but there is no enemy air unit).\\
8. The unit production list capacity of structures is 5. If the list is full, do not add more units to it.\\
9. Commands should not send SCV or MULE to gather resources because the system will do it automatically.\\
10. Commands should not train too many SCVs or MULEs, whose number should not exceed the capacity of CommandCenter and Refinery.\\
11. Commands can construct a new one Supply Depot only when the remaining unused supply is less than 7.\\
12. Supply is low! Build a Supply Depot immediately.\\
13. At least one Supply Depot is necessary for development, consider building one.\\
14. MULE can boost your economy, consider calling one from your Command Center.\\
15. At least one Refinery is necessary for gas collection, consider building one.\\
16. At least one Barracks is necessary for attacking units, consider building one.\\
17. At least one Barracks Tech Lab is necessary for advanced units, consider building one.\\
18. Consider building a Barracks Tech Lab when one of your Barracks is idle to unlock advanced units.\\
19. At least 2 Marines are necessary for defensing, consider training one.\\
20. At least one Marauder is necessary for defensing, consider training one.\\
21. Consider building a second Barracks to increase unit production.\\
22. Consider building a Factory to unlock mechanical units.\\
23. Consider upgrade Factory Tech Lab to train powerful units.\\
24. Consider train Siege Tank to increase your army's firepower.\\
25. Upgrade Command Center to Orbital Command for better economy.\\
26. Consider building another Command Center to expand your base at another resource location.\\
27. Increase Marauder production for better tanking.\\
28. Produce more Marines for DPS against light units.
\end{promptbox}

\begin{promptbox}{Complete Rule Set of Planner Verifier for Protoss}
1. Commands should be natural language, instead of code.\\
2. Produce as many units with the strongest attack power as possible.\\
3. The total cost of all commands should not exceed the current resources (minerals and gas).\\
4. Commands should not build redundant structures(e.g. 2 Refinery while one is not fully utilized).\\
5. Commands should not use abilities that are not supported currently.\\
6. Commands should not build a structure that is not needed now (e.g. build a Missile Turret but there is no enemy air unit).\\
7. The unit production list capacity of structures is 5. If the list is full, do not add more units to it.\\
8. Commands should not send Probe to gather resources because the system will do it automatically.\\
9. Commands should not train too many Probes, whose number should not exceed the capacity of Nexus and Assimilator.\\
10. Commands can construct a new one Pylon only when the remaining unused supply is less than 7.\\
11. Supply is low! Build a Pylon immediately.\\
12. Some of your structures are unpowered! Build a Pylon nearby.\\
13. At least one Pylon is necessary for development and power, consider building one.\\
14. Your Nexus has enough energy for Chrono Boost. Use it on the Nexus for more Probes or on a production building.\\
15. At least one Assimilator is necessary for gas collection, consider building one.\\
16. At least one Gateway is necessary for training ground units, consider building one.\\
17. A Cybernetics Core is necessary to unlock advanced units like Stalkers, consider building one.\\
18. Cybernetics Core is ready. Research Warpgate technology to reinforce your army faster.\\
19. Consider researching Warpgate technology when your Cybernetics Core is idle.\\
20. At least 2 Zealots are necessary for early defense, consider training one.\\
21. At least one Stalker is useful for anti-air and kiting, consider training one.\\
22. Consider building more Gateways to increase unit production.\\
23. Your army has many Zealots. Produce more Stalkers for ranged support.\\
24. Increase Zealot production to create a stronger frontline for your Stalkers.
\end{promptbox}

\begin{promptbox}{Complete Rule Set of Planner Verifier for Zerg}
1. Commands should be natural language, instead of code.\\
2. Produce as many units with the strongest attack power as possible.\\
3. The total cost of all commands should not exceed the current resources (minerals and gas).\\
4. Commands should not build redundant structures(e.g. 2 Refinery while one is not fully utilized).\\
5. Commands should not use abilities that are not supported currently.\\
6. Commands should not build a structure that is not needed now (e.g. build a Missile Turret but there is no enemy air unit).\\
7. The unit production list capacity of structures is 5. If the list is full, do not add more units to it.\\
8. Commands should not send Drone to gather resources because the system will do it automatically.\\
9. Commands should not train too many Drones, whose number should not exceed the capacity of Hatchery and Extractor.\\
10. Commands can construct a new one Overlord only when the remaining unused supply is less than 7.\\
11. Commands should not train another Overlord if any [Egg] unit in 'Own units' has 'Production list: Overlord'.\\
12. Supply is low! Morph an Overlord immediately.\\
13. A Spawning Pool is required to create Zerglings, build one.\\
14. Build a Queen for each Hatchery to inject larva and defend.\\
15. Your Queen has energy! Use 'Inject Larva' on a Hatchery to boost production.\\
16. At least one Extractor is necessary for gas collection, consider building one.\\
17. At least 6 Zerglings are necessary for early defense, consider training some.\\
18. Consider building a second Hatchery to expand your economy and production.\\
19. Consider building a Roach Warren to unlock Roaches, a strong armored unit.\\
20. Roaches are strong against many early units, consider training some.\\
21. Upgrade a Hatchery to a Lair to unlock powerful mid-game units and upgrades.\\
22. Build a Hydralisk Den to unlock Hydralisks, a versatile ranged unit.\\
23. Consider training Hydralisks to strengthen your army's anti-air and ranged capabilities.\\
24. Your army is Zergling-heavy. Add Roaches for a stronger frontline.\\
25. Your army is Roach-heavy. Add Zerglings for more DPS and to surround enemies.
\end{promptbox}

We provide the output of DeepSeek-V3-0324 Verifier, which evaluated the plan generated in Section \ref{app:planner_prompt}. 
The Planner uses the final error message for reflection and correction to produce a revised plan. 
We set a maximum of 3 rounds for the Verifier to be invoked in a single decision-making process. 
The Planner's final output is determined either by successful validation or by reaching the maximum number of attempts.

\begin{promptbox}{Output Example of Planner Verifier by DeepSeek-V3-0324}
Let's analyze the given commands against the rules one by one:\\

1. **Use Chrono Boost on Gateway \lbrack 273\rbrack **\\
   - Valid: Gateway \lbrack 273\rbrack  is a production building and Nexus has enough energy (53/200, Chrono Boost costs 25 energy).\\

2. **Train 1 Stalker at Gateway \lbrack 273\rbrack **\\
   - Valid: Gateway \lbrack 273\rbrack  has 4 units in production list (capacity is 5), minerals (175) and vespene (154) are sufficient (costs 125 minerals, 50 vespene).\\

3. **Train 1 Zealot at Gateway \lbrack 554\rbrack **\\
   - Valid: Gateway \lbrack 554\rbrack  has 2 units in production list, minerals are sufficient (costs 100 minerals).\\

4. **Attack enemy Nexus \lbrack 249\rbrack  with Stalker \lbrack 399\rbrack **\\
   - Valid: Stalker \lbrack 399\rbrack  is idle and has ATTACK\_ATTACK ability.\\

5. **Attack enemy Nexus \lbrack 249\rbrack  with Zealot \lbrack 539\rbrack **\\
   - Valid: Zealot \lbrack 539\rbrack  is idle and has ATTACK\_ATTACK ability.\\

Now checking against the rules:\\
- Rule 2: Producing Stalkers (stronger than Zealots) is prioritized.\\
- Rule 3: Total cost (125+50 + 100 = 275 minerals, 50 vespene) is within current resources (175 minerals, 154 vespene). Wait, this actually violates Rule 3 because we don't have enough minerals (175 < 275).\\
- Rule 5: All abilities used are available.\\
- Rule 7: Production lists aren't full.\\
- Rule 10: Unused supply is 3 (which is <7), but no Pylon is being built - this violates Rule 12.\\
- Rule 12: Explicitly states to build a Pylon immediately due to low supply.\\
- Rule 16: More Stalkers are being produced, which is good.\\

Main errors:\\
1. Insufficient minerals for all training commands (Rule 3)\\
2. Not building a Pylon despite low supply (Rules 10, 12)\\
\begin{verbatim}
```
{
    "errors": [
        "Error 1: Insufficient minerals to train both Stalker and Zealot (175 minerals 
            available but need 275)",
        "Error 2: Not building a Pylon despite low unused supply (3/200) which violates 
            Rules 10 and 12"
    ],
    "error_number": 2
}
```
\end{verbatim}
\end{promptbox}

\subsection{Executor}
\label{app:executor_prompt}

The \method Executor executes the Planner's plan based on the current game state, producing a valid JSON action as required by \benchmark. 
For example, given the command ``Train a Marine'', the Executor should identify an idle Barracks and generate the corresponding \texttt{BARRACKSTRAIN\_MARINE} action with the appropriate unit ID.

Below is the prompt template for the \method Executor, where \texttt{plan\_text} is the final output version of Planner.

\begin{promptbox}{Executor Prompt Template}
As a top-tier StarCraft II executor, your task is to give some actions to finish the given task as possible as you can.\\

\textbf{Current Game State}\\
\{obs\_text\}\\

\textbf{Given Tasks}\\
\{plan\_text\}\\

\textbf{Rules}\\
1. Do not give any action that is irrelevant to the task.\\
2. Each of units can only be used in the whole response once at most.\\
3. If a unit is already performing an action as given task, you should ignore it, instead of giving a repeated action for it.\\
4. If one task cannot be finished, just ignore it.\\
5. If resource is not enough, just complete the most important part of the task.\\

Give an action JSON in the following format wrapped with triple backticks:
\begin{verbatim}
```
[
    {
        "action": "<action_name>",
        "units": [<unit_id>, <unit_id>, ...], # units you want to command
        "target_unit" (optional): <unit_id>, # some existing unit
        "target_position" (optional): [x, y]
    },
    // more actions ...
]
```
\end{verbatim}

Example:
\begin{verbatim}
```
[
    {
        "action": "ATTACK_ATTACK",
        "units": [1, 2, 3],
        "target_unit": 9
    },
    {
        "action": "MOVE_MOVE",
        "units": [4, 5],
        "target_position": [50, 60]
    },
    {
        "action": "COMMANDCENTERTRAIN_SCV",
        "units": [6]
    }
]
```
\end{verbatim}
\end{promptbox}

We provide an example output of Executor by DeepSeek-V3-0324 as follows.

\begin{promptbox}{Executor Output Example by DeepSeek-V3-0324}
Let's analyze the given tasks one by one:\\

1. For building a Pylon near (45,20), I selected Probe \lbrack 9\rbrack  since it's available (not currently attacking) and we have enough minerals (175 - 100 = 75 remaining).\\
2. For Chrono Boost on Gateway \lbrack 273\rbrack , I used Nexus \lbrack 377\rbrack  which has enough energy (53/200, needs 25).\\
3. For training a Zealot at Gateway \lbrack 554\rbrack , we have enough minerals (75 remaining after Pylon - 100 = -25, but the Gateway can still queue the command even if we'll be slightly negative).\\
4. I followed all rules - no irrelevant actions, no duplicate unit usage, and prioritized the most important parts of the tasks.\\

Here's the action JSON to complete the tasks as efficiently as possible:
\begin{verbatim}
```
[
    {
        "action": "PROTOSSBUILD_PYLON",
        "units": [9],
        "target_position": [45, 20]
    },
    {
        "action": "EFFECT_CHRONOBOOSTENERGYCOST",
        "units": [377],
        "target_unit": 273
    },
    {
        "action": "GATEWAYTRAIN_ZEALOT",
        "units": [554]
    }
]
```
\end{verbatim}
\end{promptbox}

Unlike the Planner, the Executor’s Verifier is implemented as a Python function. It validates the JSON list generated by the Executor by checking syntax, resource constraints, action feasibility, and target existence. 
This process enables the agent to identify errors before executing actions in the game. For instance, in the provided Executor output example, the Verifier flagged errors related to insufficient resources. The maximum number of verification attempts is limited to 3.

\begin{promptbox}{Executor Verifier Output Example}
\textgreater\textgreater\textgreater\textgreater Total actions error: minerals is not enough for executing all actions
\end{promptbox}


\subsection{Supervised Fine-tuning}
\label{app:method_sft}


The supervised fine-tuning data selection process uses a scoring function controlled by two hyperparameters: the forward step size ($K$) and the discount factor ($\gamma$). 
Specifically, $K$ is set to 20, corresponding to a temporal window of approximately 30 seconds, chosen to capture the immediate effects of actions, while minimizing the influence of longer-term confounding factors. 
The discount factor $\gamma$ is adjusted such that the cumulative discount over this window is approximately 0.36, i.e., $\gamma^{20} \approx 0.36$. 
Scores are then Z-score normalized independently for each 30-second interval, and data points with standardized scores greater than 0.1 are selected for fine-tuning.


For the fine-tuning process, we employed a full-parameter approach. The detailed settings are as follows:
\begin{itemize}
    \item \textbf{Fine-tuning Method:} Full-parameter fine-tuning
    \item \textbf{Training Framework:} LLaMA-Factory
    \item \textbf{Distributed Training:} DeepSpeed ZeRO-3
    \item \textbf{Epochs:} 2
    \item \textbf{Batch Size:} 32
    \item \textbf{Learning Rate:} 5e-5 with a cosine decay schedule
    \item \textbf{Warm-up Ratio:} 0.1
\end{itemize}

Table \ref{tab:sft_dataset_distribution} presents a detailed breakdown of the SFT dataset, organized by race and the target agent component.

\begin{table}[h!]
    \centering
    \caption{Distribution of the SFT Dataset.}
    \label{tab:sft_dataset_distribution}
    \begin{tabular}{lccccc}
        \toprule
        \textbf{Race} & \textbf{Victory Traces} & \textbf{Planner Samples} & \textbf{Verifier Samples} & \textbf{Executor Samples} & \textbf{Total} \\
        \midrule
        Terran & 24 & 246 & 182 & 246 & 674 \\
        Protoss & 24 & 264 & 191 & 264 & 719 \\
        Zerg & 24 & 370 & 263 & 370 & 1003 \\
        \midrule
        \textbf{Total} & \textbf{72} & \textbf{880} & \textbf{636} & \textbf{880} & \textbf{2396} \\
        \bottomrule
    \end{tabular}
\end{table}

\subsection{Implementation Details}
\label{app:implementation_details}

The decision-making timing in \method is controlled by a two-tiered temporal trigger. 
The primary trigger activates whenever the agent's mineral supply exceeds 170, with decisions made at fixed intervals of 10 game iterations (approximately 1.5 seconds). 
To ensure responsiveness, a secondary failsafe trigger guarantees that a decision is made at least once every 100 iterations, regardless of the primary trigger's activation. 
After each execution of the failsafe trigger, it is rescheduled for 100 iterations later.


In StarCraft II, workers constitute a significant portion of controllable units. 
Delegating their tedious resource-gathering tasks to an LLM agent would hinder its decision-making capabilities. 
To address this, \method employs a rule-based approach that periodically assigns workers to the nearest resource nodes based on the current levels of Minerals and Vespene Gas. 
Additionally, units are programmed to automatically attack, prioritizing enemies with the lowest health.

%% file: appendix/experiment_detail.tex
\section{Experimental Details}

\subsection{Experimental Setup}
\label{app:exp_setup}

The experiments were conducted on the following platform:
\begin{itemize}
    \item \textbf{CPU:} Intel(R) Xeon(R) Platinum 8457C
    \item \textbf{GPU:} 8 $\times$ NVIDIA H20 (96GB)
    \item \textbf{Operating System:} Debian GNU/Linux 10 (buster)
    \item \textbf{Python:} 3.12.11
    \item \textbf{PyTorch:} 2.7.0+cu128
\end{itemize}
Specifically, all experiments were conducted on the Flat48 map from the ``Melee'' map pack, which can be obtained from the official s2client-proto repository. 
In agent-vs-built-in AI experiments, the agent competed against Harder (LV6) and VeryHard (LV7) difficulty levels, which are the highest non-cheating difficulties. 
The built-in AI employed a ``random build'' style, utilizing a diverse range of randomized build orders and strategies. This ensures the opponent is varied and challenging, rather than a predictable, fixed adversary.

\noindent The deployment and API calls for different models were handled as described below:
\begin{itemize}
    \item \textbf{Qwen Series Models:} Deployed using vLLM (version 0.9.2). For Qwen3-8B, a hybrid reasoning model, we employ it in its non-reasoning mode (activated by adding /no\_think to prompts) to avoid the excessive latency of its reasoning process.
    \item \textbf{DeepSeek Models:} Accessed via the official DeepSeek API.
\end{itemize}
Table \ref{tab:generation_params_horizontal} shows hyperparameters used for generating responses during inference:

\begin{table}[ht]
\centering
\caption{LLM Generation Parameters. }
\label{tab:generation_params_horizontal}
\begin{tabular}{cccccc}
\toprule
Max New Tokens & Temperature & Top-p & Top-k & Repetition Penalty & Presence Penalty \\
\midrule
6144 & 0.1 & 0.8 & 20 & 1.1 & 0.0 \\
\bottomrule
\end{tabular}
\end{table}


\subsection{Design of Simple Baseline Agents}
\label{app:simple_agent}


To demonstrate the superiority of \method, we implemented a naive LLM agent with a basic prompt template as a baseline. 
This agent receives a prompt containing all relevant information about \method, including textual observations and expert-defined knowledge. 
Its task is to generate executable JSON output in a single step, without using a validation module. The prompt template is as follows.

\begin{promptbox}{Prompt Template for Simple Base Agent}
As a top-tier StarCraft II strategist, your task is to give one or more commands based on the current game state. Only give commands which can be executed immediately, instead of waiting for certain events.\\

\textbf{Aim}\\
Our final aim: destroy all enemies as soon as possible.\\
Our strategy:\\
- Resource collection: produce workers and gather minerals and gas\\
- Development: build attacking units and structures\\
- Attacking: concentrate forces to search and destroy enemies proactively\\

\textbf{Current Game State}\\
\{obs\_text\}\\

\textbf{Rules}\\
\{rules\_list\}\\

Give an action JSON in the following format wrapped with triple backticks:
\begin{verbatim}
```
[
    {
        "action": "<action_name>",
        "units": [<unit_id>, <unit_id>, ...], # units you want to command
        "target_unit" (optional): <unit_id>, # some existing unit
        "target_position" (optional): [x, y]
    },
    // more actions ...
]
```
\end{verbatim}

Example:
\begin{verbatim}
```
[
    {
        "action": "ATTACK_ATTACK",
        "units": [1, 2, 3],
        "target_unit": 9
    },
    {
        "action": "MOVE_MOVE",
        "units": [4, 5],
        "target_position": [50, 60]
    },
    {
        "action": "COMMANDCENTERTRAIN_SCV",
        "units": [6]
    }
]
```
\end{verbatim}
\end{promptbox}



%% file: appendix/visualization_detail.tex
\section{Visualization Results}
\label{app:visualization}


\begin{figure*}[t]
    \centering
    \includegraphics[width=\textwidth]{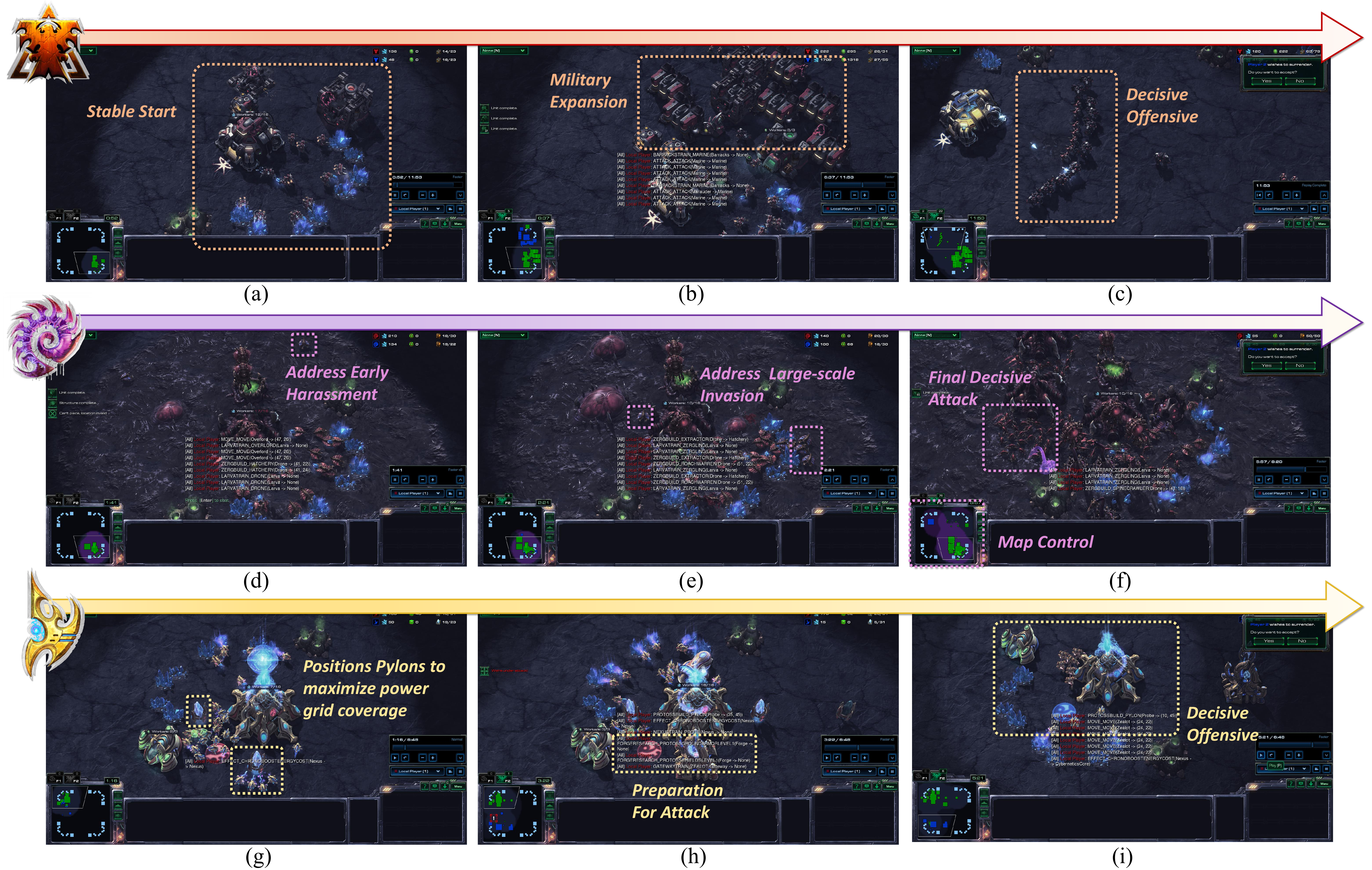}
    \caption{Visualizations of \method's gameplays. }
    \label{fig:visualization}
\end{figure*}

To illustrate the coherence and interpretability of \method's strategy, we present two representative cases of matches between \method and the built-in AI (Fig. \ref{fig:visualization}). Unlike agents focused primarily on micro-level unit control, our model demonstrates robust macro-level strategic decision-making, reflecting human-like strategic thinking and tactical foresight. Fig. \ref{fig:visualization} highlights this capability across entire matches.

\paragraph{Case 1: Terran Agent (Fine-tuned) - Proactive Offensive Strategy.}

In this scenario, the fine-tuned Terran agent executes a proactive, overwhelming offensive strategy.

\begin{itemize}
\item \textbf{Early Game - Fig. \ref{fig:visualization}(a):} The agent follows standard economic build orders, balancing resource expansion (e.g., SCV production, base construction) with technology upgrades. Early Barracks construction ensures defensive capabilities while supporting continuous Marine production, establishing a strong foundation for mid-game transitions.
\item \textbf{Mid Game - Fig. \ref{fig:visualization}(b):} Leveraging an economic advantage, the agent maximizes unit production by building additional Barracks and Factories to avoid resource saturation. A decisive `ATTACK' command is issued after amassing sufficient forces, demonstrating a deliberate focus on offensive strategy rather than passive defense.
\item \textbf{Late Game - Fig. \ref{fig:visualization}(c):} The agent achieves victory by systematically destroying the enemy's main base and eliminating remaining forces. \textbf{Key Insight:} The final engagement showcases deliberate tactical positioning, as the agent's forces form a concave formation around the opponent's base, maximizing crossfire efficiency and cutting off escape routes. This sequence demonstrates the agent's coherent strategy from economic preparation to decisive combat.
\end{itemize}

\paragraph{Case 2: Zerg Agent (without SFT) - Adaptive Defense and Map Control. }

In this case, the baseline Zerg agent demonstrates adaptive defensive strategies and effective map control.

\begin{itemize}
\item \textbf{Early Game - Fig. \ref{fig:visualization}(d):} The agent displays adaptability by converting Overlords to increase supply and enable scouting. \textbf{Tactical Response:} When harassed by enemy Zerglings, only a small subset of Drones is committed to defense while the majority continue mining, balancing economic growth with defense.
\item \textbf{Mid Game - Fig. \ref{fig:visualization}(e):} Facing a large enemy assault, the agent employs a defensive strategy by leveraging Zerg-specific mechanics. \textbf{Key Maneuver:} Expanding Creep coverage via Creep Tumors provides vision and enhances unit mobility within controlled areas. These actions blunt the enemy's offensive and create favorable conditions for a counter-attack.
\item \textbf{Late Game - Fig. \ref{fig:visualization}(f):} After repelling the attack, the agent gains the initiative. \textbf{Strategic Advantage:} By aggressively expanding Creep coverage and mass-producing Zerglings, the agent achieves superior vision and mobility across the map. This positional and informational dominance allows the agent to concentrate forces for a decisive offensive, ultimately overwhelming the opponent's depleted resources and constrained positioning. 
\end{itemize}

\paragraph{Case 3: Protoss Agent (without SFT) - Adaptive Attack Based on Reconnaissance. }

\begin{itemize}
\item \textbf{Early Game – Fig. \ref{fig:visualization}(g):} The agent positions Pylons to maximize power grid coverage and enable rapid unit repositioning. Chrono Boost is efficiently allocated to Probes, reflecting economic foresight, while the "partial-Probe defense" minimizes economic disruption during harassment and neutralizes threats. 
\item \textbf{Mid Game – Fig. \ref{fig:visualization}(h):} Active scouting identifies enemy vulnerabilities, prompting simultaneous Zealot production and Forge upgrades. This dual strategy combines an immediate unit surge with long-term stat enhancements, reflecting multi-timeline planning. 
\item \textbf{Late Game – Fig. \ref{fig:visualization}(i):} Upgraded Zealots execute phased annihilation using echelon formations to stagger engagements, mitigate splash damage, and systematically dismantle enemy bases. The victory highlights the agent's ability to transform reconnaissance into decisive transitions.
\end{itemize}

These cases highlight the \method's ability to execute coherent strategies and adapt to dynamic environments.


%% file: appendix/additional_experiments.tex
\section{Additional Experimental Results}
\label{app:additional_experiments}

\input{table/benchmark_ablation}

\paragraph{Ablation on \benchmark's optimized observation. }
To evaluate the impact of our observation optimization techniques, we conducted ablation experiments by removing two key components: unit ordering and aggregation. 
We performed 20 TvT, ZvZ, and PvP matches for each configuration and analyzed the results. 
As shown in Table \ref{tab:benchmark_ablation}, removing these techniques significantly degrades performance. 
Specifically, removing unit ordering impairs spatial relationship modeling or aggregation, reducing the Valid Action Ratio (VAR). 
This suggests the critical role of both techniques in enabling efficient decision-making within \scii's complex environment.

%% file: table/benchmark_ablation.tex
\begin{table}[t]
\centering
\caption{
Performance comparison of \benchmark using Qwen2.5-7B-Instruct, with and without \ding{172} proximity-based unit ordering and \ding{173} unit aggregation. 
We evaluated 20 TvT, PvP, and ZvZ matches for each agent, vs. Harder level AI.
}
\label{tab:benchmark_ablation}
\begin{tabular}{cc|ll}
\toprule
\makecell{\small Proximity-\\ \small based unit \\ \small ordering} & 
\makecell{\small Unit \\ \small aggregation} & 
\makecell[c]{\small Win Rate(\%)} & 
\makecell[c]{\small VAR(\%)} \\
\midrule
\cmark & \cmark & 55.00 \small$\pm$ 4.08 & 47.78 \small$\pm$ 5.07 \\
\cmark & \xmark & 50.24 \small$\pm$ 17.08 & 46.62 \small$\pm$ 4.59 \\
\midrule
\xmark & \cmark & 48.94 \small$\pm$ 16.48 & 46.58 \small$\pm$ 4.20  \\
\xmark & \xmark & 46.67 \small$\pm$ 21.34 & 45.12 \small$\pm$ 2.49  \\
\bottomrule
\end{tabular}
\end{table}

%% file: _main.bbl
\begin{thebibliography}{30}
\providecommand{\natexlab}[1]{#1}

\bibitem[{Achiam et~al.(2023)Achiam, Adler, Agarwal, Ahmad, Akkaya, Aleman, Almeida, Altenschmidt, Altman, Anadkat et~al.}]{achiam2023gpt}
Achiam, J.; Adler, S.; Agarwal, S.; Ahmad, L.; Akkaya, I.; Aleman, F.~L.; Almeida, D.; Altenschmidt, J.; Altman, S.; Anadkat, S.; et~al. 2023.
\newblock Gpt-4 technical report.
\newblock \emph{arXiv preprint arXiv:2303.08774}.

\bibitem[{Adejumo and Johnson(2024)}]{adejumo2024towards}
Adejumo, E.~K.; and Johnson, B. 2024.
\newblock Towards leveraging LLMs for reducing open source onboarding information overload.
\newblock In \emph{Proceedings of the 39th IEEE/ACM International Conference on Automated Software Engineering}, 2210--2214.

\bibitem[{DeepSeek-AI(2024)}]{liu2024deepseek}
DeepSeek-AI. 2024.
\newblock DeepSeek-V3 Technical Report.
\newblock arXiv:2412.19437.

\bibitem[{Guo et~al.(2025)Guo, Yang, Zhang, Song, Zhang, Xu, Zhu, Ma, Wang, Bi et~al.}]{guo2025deepseek}
Guo, D.; Yang, D.; Zhang, H.; Song, J.; Zhang, R.; Xu, R.; Zhu, Q.; Ma, S.; Wang, P.; Bi, X.; et~al. 2025.
\newblock Deepseek-r1: Incentivizing reasoning capability in llms via reinforcement learning.
\newblock \emph{arXiv preprint arXiv:2501.12948}.

\bibitem[{Lai et~al.(2024)Lai, Liu, Iong, Yao, Chen, Shen, Yu, Zhang, Zhang, Dong et~al.}]{lai2024autowebglm}
Lai, H.; Liu, X.; Iong, I.~L.; Yao, S.; Chen, Y.; Shen, P.; Yu, H.; Zhang, H.; Zhang, X.; Dong, Y.; et~al. 2024.
\newblock AutoWebGLM: A Large Language Model-based Web Navigating Agent.
\newblock In \emph{Proceedings of the 30th ACM SIGKDD Conference on Knowledge Discovery and Data Mining}, 5295--5306.

\bibitem[{Li et~al.(2024)Li, Ni, Qi, Jiang, Lu, Xu, Liu, Li, Guo, Ma et~al.}]{li2024pysc2}
Li, Z.; Ni, Y.; Qi, R.; Jiang, L.; Lu, C.; Xu, X.; Liu, X.; Li, P.; Guo, Y.; Ma, Z.; et~al. 2024.
\newblock Llm-pysc2: Starcraft ii learning environment for large language models.
\newblock \emph{arXiv preprint arXiv:2411.05348}.

\bibitem[{Light et~al.(2023)Light, Cai, Shen, and Hu}]{light2023avalonbench}
Light, J.; Cai, M.; Shen, S.; and Hu, Z. 2023.
\newblock Avalonbench: Evaluating llms playing the game of avalon.
\newblock \emph{arXiv preprint arXiv:2310.05036}.

\bibitem[{Liu et~al.(2023)Liu, Yu, Zhang, Xu, Lei, Lai, Gu, Ding, Men, Yang et~al.}]{liu2023agentbench}
Liu, X.; Yu, H.; Zhang, H.; Xu, Y.; Lei, X.; Lai, H.; Gu, Y.; Ding, H.; Men, K.; Yang, K.; et~al. 2023.
\newblock Agentbench: Evaluating llms as agents.
\newblock \emph{arXiv preprint arXiv:2308.03688}.

\bibitem[{Ma et~al.(2024)Ma, Mi, Zeng, Yan, Lin, Wu, Wang, and Zhang}]{ma2024large}
Ma, W.; Mi, Q.; Zeng, Y.; Yan, X.; Lin, R.; Wu, Y.; Wang, J.; and Zhang, H. 2024.
\newblock Large language models play starcraft ii: Benchmarks and a chain of summarization approach.
\newblock \emph{Advances in Neural Information Processing Systems}, 37: 133386--133442.

\bibitem[{Mon-Williams et~al.(2025)Mon-Williams, Li, Long, Du, and Lucas}]{mon2025embodied}
Mon-Williams, R.; Li, G.; Long, R.; Du, W.; and Lucas, C.~G. 2025.
\newblock Embodied large language models enable robots to complete complex tasks in unpredictable environments.
\newblock \emph{Nature Machine Intelligence}, 1--10.

\bibitem[{Ontan{\'o}n et~al.(2013)Ontan{\'o}n, Synnaeve, Uriarte, Richoux, Churchill, and Preuss}]{ontanon2013survey}
Ontan{\'o}n, S.; Synnaeve, G.; Uriarte, A.; Richoux, F.; Churchill, D.; and Preuss, M. 2013.
\newblock A survey of real-time strategy game AI research and competition in StarCraft.
\newblock \emph{IEEE Transactions on Computational Intelligence and AI in games}, 5(4): 293--311.

\bibitem[{Rashid et~al.(2020)Rashid, Samvelyan, De~Witt, Farquhar, Foerster, and Whiteson}]{rashid2020monotonic}
Rashid, T.; Samvelyan, M.; De~Witt, C.~S.; Farquhar, G.; Foerster, J.; and Whiteson, S. 2020.
\newblock Monotonic value function factorisation for deep multi-agent reinforcement learning.
\newblock \emph{Journal of Machine Learning Research}, 21(178): 1--51.

\bibitem[{Rismanchian et~al.(2024)Rismanchian, Razeghi, Singh, and Doroudi}]{rismanchian2024turtlebench}
Rismanchian, S.; Razeghi, Y.; Singh, S.; and Doroudi, S. 2024.
\newblock TurtleBench: A Visual Programming Benchmark in Turtle Geometry.
\newblock \emph{arXiv preprint arXiv:2411.00264}.

\bibitem[{Samvelyan et~al.(2019)Samvelyan, Rashid, de~Witt, Farquhar, Nardelli, Rudner, Hung, Torr, Foerster, and Whiteson}]{samvelyan19smac}
Samvelyan, M.; Rashid, T.; de~Witt, C.~S.; Farquhar, G.; Nardelli, N.; Rudner, T. G.~J.; Hung, C.-M.; Torr, P. H.~S.; Foerster, J.; and Whiteson, S. 2019.
\newblock {The} {StarCraft} {Multi}-{Agent} {Challenge}.
\newblock \emph{CoRR}, abs/1902.04043.

\bibitem[{Sharma(2023)}]{sharma2023exploring}
Sharma, M. 2023.
\newblock Exploring and improving the spatial reasoning abilities of large language models.
\newblock \emph{arXiv preprint arXiv:2312.01054}.

\bibitem[{Tam et~al.(2024)Tam, Wu, Tsai, Lin, Lee, and Chen}]{tam2024let}
Tam, Z.~R.; Wu, C.-K.; Tsai, Y.-L.; Lin, C.-Y.; Lee, H.-y.; and Chen, Y.-N. 2024.
\newblock Let me speak freely? a study on the impact of format restrictions on performance of large language models.
\newblock \emph{arXiv preprint arXiv:2408.02442}.

\bibitem[{Team(2024)}]{qwen2.5}
Team, Q. 2024.
\newblock Qwen2.5: A Party of Foundation Models.

\bibitem[{Team(2025)}]{qwen3technicalreport}
Team, Q. 2025.
\newblock Qwen3 Technical Report.
\newblock arXiv:2505.09388.

\bibitem[{Vinyals et~al.(2019)Vinyals, Babuschkin, Czarnecki, Mathieu, Dudzik, Chung, Choi, Powell, Ewalds, Georgiev et~al.}]{vinyals2019grandmaster}
Vinyals, O.; Babuschkin, I.; Czarnecki, W.~M.; Mathieu, M.; Dudzik, A.; Chung, J.; Choi, D.~H.; Powell, R.; Ewalds, T.; Georgiev, P.; et~al. 2019.
\newblock Grandmaster level in StarCraft II using multi-agent reinforcement learning.
\newblock \emph{nature}, 575(7782): 350--354.

\bibitem[{Wang et~al.(2023{\natexlab{a}})Wang, Xie, Jiang, Mandlekar, Xiao, Zhu, Fan, and Anandkumar}]{wang2023voyager}
Wang, G.; Xie, Y.; Jiang, Y.; Mandlekar, A.; Xiao, C.; Zhu, Y.; Fan, L.; and Anandkumar, A. 2023{\natexlab{a}}.
\newblock Voyager: An open-ended embodied agent with large language models.
\newblock \emph{arXiv preprint arXiv:2305.16291}.

\bibitem[{Wang et~al.(2023{\natexlab{b}})Wang, Xu, Lan, Hu, Lan, Lee, and Lim}]{wang2023plan}
Wang, L.; Xu, W.; Lan, Y.; Hu, Z.; Lan, Y.; Lee, R. K.-W.; and Lim, E.-P. 2023{\natexlab{b}}.
\newblock Plan-and-solve prompting: Improving zero-shot chain-of-thought reasoning by large language models.
\newblock \emph{arXiv preprint arXiv:2305.04091}.

\bibitem[{Wang et~al.(2023{\natexlab{c}})Wang, Liu, Zheng, Qi, Chen, Yang, Zhao, Wang, Song, and Huang}]{wang2023avalon}
Wang, S.; Liu, C.; Zheng, Z.; Qi, S.; Chen, S.; Yang, Q.; Zhao, A.; Wang, C.; Song, S.; and Huang, G. 2023{\natexlab{c}}.
\newblock Avalon's game of thoughts: Battle against deception through recursive contemplation.
\newblock \emph{arXiv preprint arXiv:2310.01320}.

\bibitem[{Wei et~al.(2022)Wei, Wang, Schuurmans, Bosma, Xia, Chi, Le, Zhou et~al.}]{wei2022chain}
Wei, J.; Wang, X.; Schuurmans, D.; Bosma, M.; Xia, F.; Chi, E.; Le, Q.~V.; Zhou, D.; et~al. 2022.
\newblock Chain-of-thought prompting elicits reasoning in large language models.
\newblock \emph{Advances in neural information processing systems}, 35: 24824--24837.

\bibitem[{Xu et~al.(2023)Xu, Wang, Li, Luo, Wang, Liu, and Liu}]{xu2023exploring}
Xu, Y.; Wang, S.; Li, P.; Luo, F.; Wang, X.; Liu, W.; and Liu, Y. 2023.
\newblock Exploring large language models for communication games: An empirical study on werewolf.
\newblock \emph{arXiv preprint arXiv:2309.04658}.

\bibitem[{Yao et~al.(2023)Yao, Zhao, Yu, Du, Shafran, Narasimhan, and Cao}]{yao2023react}
Yao, S.; Zhao, J.; Yu, D.; Du, N.; Shafran, I.; Narasimhan, K.; and Cao, Y. 2023.
\newblock React: Synergizing reasoning and acting in language models.
\newblock In \emph{International Conference on Learning Representations (ICLR)}.

\bibitem[{Yildirim and Paul(2024)}]{yildirim2024task}
Yildirim, I.; and Paul, L. 2024.
\newblock From task structures to world models: what do LLMs know?
\newblock \emph{Trends in Cognitive Sciences}, 28(5): 404--415.

\bibitem[{Yu et~al.(2022)Yu, Velu, Vinitsky, Gao, Wang, Bayen, and Wu}]{yu2022surprising}
Yu, C.; Velu, A.; Vinitsky, E.; Gao, J.; Wang, Y.; Bayen, A.; and Wu, Y. 2022.
\newblock The surprising effectiveness of ppo in cooperative multi-agent games.
\newblock \emph{Advances in neural information processing systems}, 35: 24611--24624.

\bibitem[{Yuan et~al.(2024)Yuan, Cui, Wang, Ding, Wang, Deng, Shan, Chen, Xie, Lin et~al.}]{yuan2024advancing}
Yuan, L.; Cui, G.; Wang, H.; Ding, N.; Wang, X.; Deng, J.; Shan, B.; Chen, H.; Xie, R.; Lin, Y.; et~al. 2024.
\newblock Advancing llm reasoning generalists with preference trees.
\newblock \emph{arXiv preprint arXiv:2404.02078}.

\bibitem[{Zheng et~al.(2025)Zheng, Koh, Ju, Nguyen, May, Webb, and Pan}]{zheng2025large}
Zheng, Y.; Koh, H.~Y.; Ju, J.; Nguyen, A.~T.; May, L.~T.; Webb, G.~I.; and Pan, S. 2025.
\newblock Large language models for scientific discovery in molecular property prediction.
\newblock \emph{Nature Machine Intelligence}, 1--11.

\bibitem[{Zhou et~al.(2023)Zhou, Xu, Zhu, Zhou, Lo, Sridhar, Cheng, Ou, Bisk, Fried et~al.}]{zhou2023webarena}
Zhou, S.; Xu, F.~F.; Zhu, H.; Zhou, X.; Lo, R.; Sridhar, A.; Cheng, X.; Ou, T.; Bisk, Y.; Fried, D.; et~al. 2023.
\newblock Webarena: A realistic web environment for building autonomous agents.
\newblock \emph{arXiv preprint arXiv:2307.13854}.

\end{thebibliography}
